\documentclass[letterpaper,journal]{IEEEtran}
\usepackage[T1]{fontenc}
\usepackage[utf8]{inputenc}
\usepackage{amsmath,amssymb,amsfonts}
\usepackage{graphicx}
\usepackage{booktabs,array,multirow,hhline}
\usepackage[caption=false,font=footnotesize]{subfig}
\usepackage{placeins}
\usepackage[table]{xcolor}
\usepackage{cite}
\usepackage{url}
\usepackage{microtype}
\usepackage[hidelinks]{hyperref}
\usepackage{etoolbox}
\graphicspath{{figures/}}
\newcommand{\mH}{\mathbf{H}}
\newcommand{\mW}{\mathbf{W}}
\newcommand{\mX}{\mathbf{X}}
\newcommand{\va}{\mathbf{a}}
\newcommand{\ve}{\mathbf{e}}
\newcommand{\vr}{\mathbf{r}}
\newcommand{\vw}{\mathbf{w}}
\newcommand{\vz}{\mathbf{z}}
\newcommand{\balpha}{\boldsymbol{\alpha}}

\begin{document}
\bstctlcite{IEEEcontrol}

\title{MARR: Module-Adaptive Residual Reconstruction for Low-Bit Post-Training Quantization}
\author{Le~Su, Xing~Luo, and Zhi~Jin, \textit{Senior Member, IEEE}%
\thanks{This work was supported by the National Natural Science Foundation of China (Grants 62671633 and U24A20251) and the Shenzhen Science and Technology Program (Grants QNXMB20250701093543046 and JCYJ20230807111107015).}%
\thanks{Le Su is with the School of Intelligent Systems Engineering, Shenzhen Campus of Sun Yat-sen University, Shenzhen, Guangdong 518107, China (e-mail: sule@mail2.sysu.edu.cn).}%
\thanks{Zhi Jin is with the School of Intelligent Systems Engineering, Shenzhen Campus of Sun Yat-sen University, Shenzhen, Guangdong 518107, China, also with the Guangdong Provincial Key Laboratory of Fire Science and Intelligent Emergency Technology, Shenzhen, Guangdong 518107, China, and also with the Guangdong Provincial Key Laboratory of Robotics and Digital Intelligent Manufacturing Technology, Guangzhou, Guangdong 510535, China (e-mail: jinzh26@mail.sysu.edu.cn).}%
\thanks{Xing Luo is with the Department of Frontier Research, Peng Cheng Laboratory, Shenzhen, Guangdong 518055, China (e-mail: luox@pcl.ac.cn).}%
\thanks{Corresponding author: Zhi Jin.}}
\markboth{IEEE Transactions on Circuits and Systems for Video Technology}%
{Su \MakeLowercase{\textit{et al.}}: Module-Adaptive Residual Reconstruction}
\maketitle 

\begin{abstract}
Recently, residual reconstruction-based model quantization methods have achieved promising performance in low-bit Post-Training Quantization (PTQ) by introducing cross-module residuals to reduce error accumulated from previous modules.
However, these residuals may also introduce additional bias arising from the Hessian-Approximation (HA) assumption, resulting their two-sided effect and potentially leading to suboptimal quantization performance.
Therefore, in this work, we propose a residual regulation strategy to mitigate the HA bias associated with residual strength, while preserving accumulated-error correction. Furthermore, we observe that this two-sided effect is module-dependent, we futher propose a Module-Adaptive Residual Reconstruction (MARR), which assigns a module-specific scaling coefficient to adaptively balance accumulated-error correction and residual-related HA bias for each module.
To avoid exhaustive per-module coefficient search and obtain a stable coefficient estimate, we design an Proportional-Integral-Derivative (PID)-based update strategy that employs reconstruction error as feedback to progressively refine this coefficient.  
Experiments on several typical Large Language Models (LLMs) and Vision Transformers (ViTs) demonstrate the effectiveness of MARR under low-bit quantization ($\leq$4-bit), achieving up to 20.2\% perplexity reduction on LLMs and up to 4.6\% accuracy gains on ViTs over the residual reconstruction baselines.
Code will be made publicly available upon acceptance.
\end{abstract}


\section{Introduction}
\label{sec:introduction}

Large Language Models (LLMs)~\cite{touvron2023llama,yang2025qwen3} and Vision Transformers (ViTs)~\cite{liu2021swin,dosovitskiy2020image} have achieved remarkable progress across diverse tasks. However, their large model sizes and dense operations incur substantial memory, bandwidth, and inference overhead.
To improve deployment efficiency, existing studies either compress pre-trained models through pruning~\cite{sun2024simple,cheng2024survey}, knowledge distillation~\cite{zhao2022decoupled,gou2021knowledge}, low-rank decomposition~\cite{yang2023self}, and quantization~\cite{gong2025pushing,zheng2026first}, or directly design lightweight architectures~\cite{liu2024lightweight} and efficient operators~\cite{qiu2023mb,jin2025mb}.
Among them, model quantization is deployment-friendly since it reduces the numerical precision of model weights or activations while largely preserving the original architecture. Therefore, it is a plug-and-play compression technique compatible with existing inference pipelines and hardware kernels~\cite{frantar2022gptq}.
Moreover, pushing quantization below conventional 8-bit settings is increasingly important, as low-bit quantization can further reduce memory and bandwidth costs, allowing for larger models, longer sequences, or higher throughput under the same hardware budget~\cite{gong2025pushing}.


Existing quantization methods are broadly divided into Quantization-Aware Training (QAT)~\cite{liu2024spinquant} and Post-Training Quantization (PTQ)~\cite{li2021brecq,wei2022qdrop}. 
Compared with QAT, PTQ avoids additional training or finetuning, making it more practical for large-scale pre-trained models. 
Current PTQ methods can be roughly grouped into quantization parameter-based methods~\cite{zhong2023s}, redistribution-based methods~\cite{xiao2023smoothquant,ashkboos2024quarot}, and reconstruction-based methods~\cite{frantar2022gptq,li2025gptaq}. 
Among them, reconstruction-based methods have attracted increasing attention for large models, since they efficiently reduce quantization error by adjusting weights while maintaining low quantization overhead through closed-form updates.
A representative method is GPTQ~\cite{frantar2022gptq}, which extends Optimal Brain Surgeon (OBS)~\cite{hassibi1993optimal} and Optimal Brain Quantization (OBQ)~\cite{frantar2022optimal} by applying Hessian-based column-wise updates for each module. 
However, under low-bit quantization, quantization errors from preceding modules accumulate across the network, making local-only basic reconstruction insufficient.
To address this issue, recent residual reconstruction-based methods (GPTAQ~\cite{li2025gptaq} and ResComp~\cite{li2026rethinking}) account for cross-module activation dependencies by introducing a residual between Full-Precision (FP) and Quantized (Quant) activations, which improves low-bit quantization performance.

This cross-module residual is beneficial for reducing accumulated quantization errors; however, it may also introduce additional Hessian-Approximation (HA) bias.
Such a two-sided effect makes the overall reconstruction suboptimal when overly strong residual correction is applied.
To address this issue, in this work, we propose a residual regulation strategy that controls the strength of residual correction to mitigate HA-induced bias while preserving effective compensation for accumulated quantization errors.
Moreover, the desired modulation is inherently module-dependent: the same residual strength can lead to different reconstruction behaviors across modules.
Accordingly, we propose Module-Adaptive Residual Reconstruction (MARR), which
assigns a module-specific scaling coefficient to the residual term of each
module.
By directly scaling the residual, this coefficient explicitly controls how much cross-module information is injected into each module, balancing accumulated-error correction against HA bias.
In terms of coefficient estimation efficiency, we formulate residual-strength control as a closed-loop feedback problem, where module-level reconstruction error serves as the observable feedback signal for a Proportional-Integral-Derivative (PID)-based adaptive update, avoiding exhaustive coefficient search.

Our contributions are summarized as follows:
\begin{itemize}
    \item We propose MARR, a new framework to address the suboptimal performance of residual reconstruction-based methods under low-bit PTQ.

    \item We develop a module-wise residual regulation strategy to balance accumulated-error compensation and HA-induced bias, and design a PID-based feedback mechanism to adaptively adjust residual correction strength based on reconstruction-error feedback.

    \item Experiments on LLMs and ViTs demonstrate substantial gains over residual reconstruction methods.
\end{itemize}

\section{Related Work}
\label{sec:related_work}

\subsection{Low-Bit Post-Training Quantization}
\label{sec:reconstruction_ptq}
PTQ compresses pretrained models into low-precision representations without additional retraining, making it practical for large-scale models~\cite{liu2025low,lin2026quantization,wu2022efficient}.
Existing PTQ methods improve low-bit robustness from several perspectives.
Quantization-parameter-based methods~\cite{zhong2023s} optimize clipping ranges, scaling factors, zero-points, or group-wise configurations to reduce discretization error.
Redistribution-based methods reshape weight or activation distributions through smoothing~\cite{xiao2023smoothquant}, re-parameterization~\cite{li2023repq}, rotation~\cite{ashkboos2024quarot}, or outlier migration~\cite{lee2024owq}, making models more amenable to low-bit quantization.

Reconstruction-based PTQ minimizes module-wise output discrepancies through local MSE objectives and lightweight weight compensation~\cite{nagel2020up}. GPTQ~\cite{frantar2022gptq} adopts the second-order compensation principle of OBS~\cite{hassibi1993optimal} for sequential weight reconstruction. Subsequent methods further improve reconstruction from different perspectives. FOEM~\cite{zheng2026first} addresses first-order errors, GPTAQ~\cite{li2025gptaq} compensates for propagated activation mismatch, ResComp~\cite{li2026rethinking} models compensation-aware residual errors, and CoreQ~\cite{cha2026coreq} regularizes activation-mismatch correction. Despite these advances, the interaction between residual information and HA underlying the local MSE objective remains insufficiently explored. To address this issue, we analyze the two-sided effect of residual reconstruction under HA and propose MARR for module-adaptive residual control.

\subsection{PID-Based Feedback Learning}
\label{sec:pid_feedback_learning}

PID-based learning introduces closed-loop feedback into deep learning, where the response of the current system state is observed and fed back to adapt a controllable variable.
Existing studies have applied this principle to different control targets.
For example, An et al.~\cite{an2018pid} exploited the current, accumulated, and changing gradients in a PID-inspired optimizer to update network parameters and alleviate overshoot during training.
Sharma et al.~\cite{sharmaglobally} further formulated PID controller parameters as learnable control policies and optimized them through policy-gradient methods.
Beyond parameter optimization, Ma et al.~\cite{ma2021pid,ma2022meta} integrated PID control with attention learning to dynamically regulate feature learning for image denoising, while Chen et al.~\cite{chen2024pid} employed PID feedback to correct internal network states during inference for improving model robustness.
Despite their different control targets, these methods share a common closed-loop mechanism in which an observable response is measured and fed back to guide subsequent adjustment.
Inspired by this principle, we adopt PID-based feedback control to adaptively estimate the module-specific residual scaling coefficient for balancing the two-sided effect of residual reconstruction.

\section{The Two-Sided Effect of Residual Reconstruction}
\label{sec:two_sided_effect}

In this section, we investigate the two-sided effect of residual reconstruction. We first introduce the HA underlying the local MSE surrogate, then review basic and residual reconstruction, and finally analyze the residual-related contributions to the HA-induced approximation discrepancy.

\subsection{Hessian Approximation}
\label{sec:hessian_approximation}

The local MSE objective in reconstruction-based PTQ can be viewed as a tractable proxy of the Hessian-based local task-loss objective.
A key step is to simplify the second-order derivative of the task loss with respect to the module activation through HA~\cite{nagel2020up}.
Specifically, the HA can be expressed as
\begin{equation}
\mH_{\vz}
=
\nabla_{\vz}^2 \mathcal{L}_{\mathrm{task}}
\approx
\operatorname{diag}(c_1,\ldots,c_k)
=
\bar{\mH}_{\vz} ,
\label{eq:adaround_hessian_approx}
\end{equation}
where $\mH_{\vz}\in\mathbb{R}^{k\times k}$ denotes the Hessian of the task loss with respect to the module output $\vz\in\mathbb{R}^{1\times k}$ over $k$ calibration tokens~\cite{gong2025pushing}, and $\bar{\mH}_{\vz}$ is its approximated counterpart under HA. Here, $c_i$ is treated as the $i$-th sample-independent constant.
Following AdaRound~\cite{nagel2020up}, the output Hessian is approximated as a diagonal matrix whose diagonal entries are further treated as sample-independent constants, leading to a local MSE metric.

\subsection{Basic and Residual Reconstruction}
\label{sec:basic_residual_reconstruction}

Reconstruction-based PTQ reduces quantization-induced output distortion by compensating the quantized weights with closed-form weight updates. Following GPTAQ~\cite{li2025gptaq}, we adopt the row-vector notation, where bold uppercase and lowercase letters denote matrices and vectors, respectively. For a linear module, let $\mW\in\mathbb{R}^{p\times n}$ denote the FP weight matrix, where $p$ and $n$ are the numbers of output and input channels, respectively. Let $\vw\in\mathbb{R}^{1\times n}$ denote a row of $\mW$, i.e., the weights of one output channel, and let $w_q$ denote its $q$-th element. The corresponding quantized weights are $\hat{\mW}=\operatorname{quant}(\mW)\in\mathbb{R}^{p\times n}$ and $\hat{\vw}=\operatorname{quant}(\vw)\in\mathbb{R}^{1\times n}$. Moreover, $\mX\in\mathbb{R}^{n\times k}$ and $\hat{\mX}\in\mathbb{R}^{n\times k}$ denote the FP calibration input and the corresponding input produced by the already quantized preceding modules, respectively, where $k$ is the total number of calibration tokens. Since all rows of $\mW$ share the same input and can be reconstructed in parallel, the following derivation is presented for a single row $\vw$.

\textbf{Basic Reconstruction.}
Built upon the reconstruction principle of OBS~\cite{hassibi1993optimal} and OBQ~\cite{frantar2022optimal}, GPTQ~\cite{frantar2022gptq} adjusts the remaining weights to compensate for the perturbation introduced by quantizing the current weight coordinate. Its reconstruction problem is formulated as
\begin{equation}
\min_{\Delta \vw}
\left\|\Delta \vw \hat{\mX}\right\|_F^2,
\qquad
\mathrm{s.t.}\quad
\Delta \vw = \hat{\vw} - \vw ,
\label{eq:basic_reconstruction}
\end{equation}
where $\Delta \vw\in\mathbb{R}^{1\times n}$ denotes the weight compensation, and the constraint $\Delta \vw = \hat{\vw} - \vw$ ensures that the compensated weights equal their quantized values. Let $\mH=\hat{\mX}\hat{\mX}^{\top}\in\mathbb{R}^{n\times n}$ denote the Hessian matrix. For the quantization of $w_q$, solving Eq.~\eqref{eq:basic_reconstruction} yields the closed-form update
\begin{equation}
\Delta \vw_{\mathrm{basic}}
=
\frac{\hat{w}_q-w_q}{\mH^{-1}_{qq}}
\mH^{-1}_{q,:},
\label{eq:basic_update}
\end{equation}
$\Delta \vw_{\mathrm{basic}}$ denotes the weight update obtained by basic reconstruction, while $\mH^{-1}_{qq}$ and $\mH^{-1}_{q,:}$ denote the $q$-th diagonal element and the $q$-th row of the inverse matrix $\mH^{-1}$, respectively.

The update in Eq.~\eqref{eq:basic_update} minimizes the discrepancy between $(\vw+\Delta \vw)\hat{\mX}$ and $\vw\hat{\mX}$ by adjusting the remaining weights.
However, since $\hat{\mX}$ already contains errors propagated from preceding quantized modules, preserving $\vw\hat{\mX}$ does not necessarily preserve the FP output $\vw\mX$.
Consequently, the uncorrected input mismatch may lead to accumulated quantization errors across modules~\cite{li2025gptaq}.

\textbf{Residual Reconstruction.}
Residual reconstruction addresses this limitation by explicitly incorporating the mismatch between the FP and quantized activation paths. A representative method, GPTAQ~\cite{li2025gptaq}, defines the propagated residual as $\vr=\vw\mX-\vw\hat{\mX}=\vw(\mX-\hat{\mX})\in\mathbb{R}^{1\times k}$. Accordingly, the discrepancy with respect to the FP output can be written as $\Delta \vz=(\vw+\Delta \vw)\hat{\mX}-\vw\mX=\Delta \vw\hat{\mX}-\vr$. Its reconstruction objective becomes
\begin{equation}
\min_{\Delta \vw}
\left\|\Delta \vw\hat{\mX}-\vr\right\|_F^2,
\qquad
\mathrm{s.t.}\quad
\Delta \vw = \hat{\vw}-\vw .
\label{eq:residual_reconstruction}
\end{equation}

Under the same coordinate-wise reconstruction procedure, Eq.~\eqref{eq:residual_reconstruction} admits the closed-form update
\begin{equation}
\Delta \vw_{\mathrm{res}}
=
\underbrace{
\frac{\hat{w}_q-w_q}{\mH^{-1}_{qq}}
\mH^{-1}_{q,:}
}_{\text{basic reconstruction}}
+
\underbrace{
\vr\hat{\mX}^{\top}\mH^{-1}_{-q}
}_{\text{residual correction}},
\label{eq:residual_update}
\end{equation}
where $\Delta \vw_{\mathrm{res}}$ denotes the weight update obtained by residual reconstruction, and $\mH^{-1}_{-q} = \mH^{-1} - \mH^{-1}_{:,q}\mH^{-1}_{q,:}/\mH^{-1}_{qq}$ denotes the inverse matrix after eliminating the $q$-th coordinate.

By incorporating the propagated FP--Quant activation mismatch, GPTAQ provides cross-module error compensation to reduce accumulated quantization errors.
This cross-module correction is particularly beneficial under aggressive low-bit quantization, where activation mismatch becomes more pronounced.
Nevertheless, the reconstruction objective in Eq.~\eqref{eq:residual_reconstruction} remains a local MSE-based surrogate of the task-loss objective.

\subsection{Hessian-Approximation Bias in Residual Reconstruction}
\label{sec:ha_bias}

With HA and the two reconstruction formulations established, we now analyze how the propagated residual enters the discrepancy between the Hessian-based task-loss objective and its local MSE surrogate.

\textbf{Residual Reconstruction under HA.}
Under low-bit quantization, the input of the current module is also perturbed by preceding quantized modules.
Since the quantized output is $(\vw+\Delta \vw)\hat{\mX}$ while the FP reference is $\vw\mX$.
Accordingly, the local residual reconstruction objective can be written as
\begin{equation}
\begin{aligned}
\mathcal{J}
&=
\mathbb{E}\!\left[
\Delta \vz \mH_{\vz} \Delta \vz^\top
\right] \\
&=
\mathbb{E}\!\left[
(\Delta \vw\hat{\mX}-\vr)
\mH_{\vz}
(\Delta \vw\hat{\mX}-\vr)^\top
\right] \\
&\overset{\mathrm{HA}}{\approx}
\mathbb{E}\!\left[
\left\|
\Delta \vw\hat{\mX}-\vr
\right\|_F^2
\right].
\end{aligned}
\label{eq:residual_hessian_to_mse}
\end{equation}
where $\mathbb{E}[\cdot]$ denotes the expectation over the calibration data.
The residual term $\vr$ captures the FP--Quant mismatch propagated from preceding quantized modules, allowing the local reconstruction objective to account for cross-module dependency.

\textbf{HA Bias in Residual.}
Although the residual term provides cross-module correction, Eq.~\eqref{eq:residual_hessian_to_mse} still relies on the HA in Eq.~\eqref{eq:adaround_hessian_approx}.
Let
$\Delta \mH_{\vz} = \mH_{\vz}-\bar{\mH}_{\vz}$
denote the gap between the true task-loss Hessian and its approximated counterpart.
We define the HA-induced objective discrepancy as
\[
\mathcal{E}(\vr)
=
\left|
\mathbb{E}\!\left[
(\Delta \vw\hat{\mX}-\vr)
\Delta \mH_{\vz}
(\Delta \vw\hat{\mX}-\vr)^\top
\right]
\right|.
\]

For compactness, let $\va=\Delta \vw\hat{\mX}$ denote the weight-induced output perturbation.
Since both $\mH_{\vz}$ and $\bar{\mH}_{\vz}$ are symmetric curvature matrices, $\Delta \mH_{\vz}$ is also symmetric, yielding
$\va \Delta \mH_{\vz} \vr^\top = \vr \Delta \mH_{\vz} \va^\top$.
Therefore, the HA-induced objective discrepancy can be exactly decomposed as
\begin{equation}
\begin{aligned}
\mathcal{E}(\vr)
&=
\left|
\mathbb{E}\!\left[
(\va-\vr)
\Delta \mH_{\vz}
(\va-\vr)^\top
\right]
\right| \\
&=
\left|
\mathbb{E}\!\left[
\va \Delta \mH_{\vz} \va^\top
-2\va \Delta \mH_{\vz} \vr^\top
+\vr \Delta \mH_{\vz} \vr^\top
\right]
\right|.
\end{aligned}
\label{eq:residual_hessian_error_decomposition}
\end{equation}

To separate the magnitude of each contribution without relying on their signs, we further apply the triangle inequality~\cite{billingsley1995probability} together with the standard relation
$|\mathbb{E}[u]|\leq \mathbb{E}[|u|]$.
Accordingly, Eq.~\eqref{eq:residual_hessian_error_decomposition} can be bounded as
\begin{equation}
\begin{aligned}
\mathcal{E}(\vr)
\leq {} &
\underbrace{
\mathbb{E}\!\left[
\left|
\va \Delta \mH_{\vz} \va^\top
\right|
\right]
}_{\text{weight-perturbation error}} \\
&+\underbrace{
2\mathbb{E}\!\left[
\left|
\va \Delta \mH_{\vz} \vr^\top
\right|
\right]
+
\mathbb{E}\!\left[
\left|
\vr \Delta \mH_{\vz} \vr^\top
\right|
\right]
}_{\text{residual-related error}} .
\end{aligned}
\label{eq:residual_hessian_error_bound}
\end{equation}

As shown in Eq.~\eqref{eq:residual_hessian_error_bound}, the upper bound of the HA-induced objective discrepancy consists of one weight-perturbation term and two residual-related terms.
The first term depends only on the weight-induced perturbation $\va=\Delta \vw\hat{\mX}$ and is also present in basic reconstruction.
The latter two terms explicitly depend on the propagated residual $\vr$, and thus constitute the residual-related error in the upper bound of the HA-induced objective discrepancy.

This reveals the two-sided effect of residual reconstruction, which provides cross-module correction for accumulated quantization errors while also introducing additional approximation-error components under the Hessian-to-MSE surrogate.

\section{Module-Adaptive Residual Reconstruction}
\label{sec:marr}
Building on the two-sided effect,
we propose MARR to adaptively control
the propagated residual contribution.
Specifically, we first introduce residual scaling to balance this
effect, then extend the scaling coefficient to individual modules, and finally
develop a PID-based feedback strategy for efficient coefficient estimation.

\subsection{Balancing the Two-Sided Effect of Residual}
\label{sec:residual_scaling}

The analysis in Section~\ref{sec:two_sided_effect} reveals the two-sided effect of residual reconstruction.
Therefore, instead of always applying the full residual correction, we introduce a scaling coefficient $\balpha$ to accurately control the residual contribution in the local reconstruction objective.

Specifically, the scaled-residual reconstruction problem is formulated as
\begin{equation}
\min_{\Delta \vw}
\left\|
\Delta \vw\hat{\mX}-\balpha \vr
\right\|_F^2,
\qquad
\mathrm{s.t.}\quad
\Delta \vw=\hat{\vw}-\vw .
\label{eq:scaled_residual_mse}
\end{equation}

Here, $\balpha$ determines how strongly the propagated residual is incorporated into the reconstruction.

\begin{figure*}[!t]
  \centering
  \includegraphics[width=\linewidth]{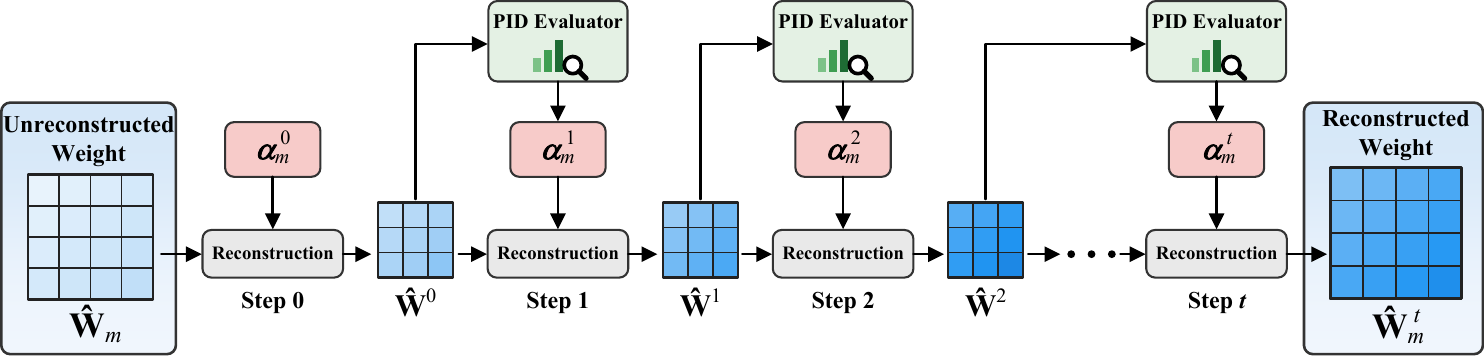}
  \caption{Overview of the PID-based adaptive update strategy for the
  module-specific residual scaling coefficient $\balpha_m$.}
  \label{fig:pid_overview}
\end{figure*}

To clarify how $\balpha$ affects the reconstruction update, we derive the closed-form solution of Eq.~\eqref{eq:scaled_residual_mse}.
When quantizing the $q$-th weight, we impose the constraint $\Delta \vw\ve_q^{\top}=\hat{w}_q-w_q$.
The corresponding Lagrangian is
\begin{equation}
\mathcal{L}(\Delta \vw,\lambda)
=
\left\|
\Delta \vw\hat{\mX}-\balpha \vr
\right\|_F^2
+
\lambda
\left(
\Delta \vw\ve_q^{\top}+w_q-\hat{w}_q
\right),
\label{eq:scaled_residual_lagrangian}
\end{equation}
where $\lambda$ is the Lagrange multiplier, and $\ve_q\in\mathbb{R}^{1\times n}$ denotes the standard basis row vector whose $q$-th element is one and all other elements are zero.

The optimal solution is obtained by taking the derivatives of the Lagrangian:
\begin{equation}
\begin{cases}
\displaystyle
\frac{\partial\mathcal{L}}{\partial\Delta \vw}
=
2\Delta \vw\mH
-2\balpha\vr\hat{\mX}^{\top}
+\lambda\ve_q,\\[8pt]
\displaystyle
\frac{\partial\mathcal{L}}{\partial\lambda}
=
\Delta \vw\ve_q^{\top}+w_q-\hat{w}_q,
\end{cases}
\label{eq:scaled_residual_kkt}
\end{equation}
where $\mH=\hat{\mX}\hat{\mX}^{\top}$.

Setting these derivatives to zero, we obtain
\begin{equation}
\Delta \vw
=
\balpha \vr\hat{\mX}^{\top}\mH^{-1}
-
\frac{\lambda}{2}\mH^{-1}_{q,:}.
\label{eq:scaled_residual_delta_w}
\end{equation}

Applying the coordinate constraint, the Lagrange multiplier is obtained as
$\lambda/2
=
\bigl(\balpha \vr\hat{\mX}^{\top}\mH^{-1}_{:,q}-(\hat{w}_q-w_q)\bigr)/\mH^{-1}_{qq}$.
Substituting this result into Eq.~\eqref{eq:scaled_residual_delta_w} yields the closed-form weight update
\begin{equation}
\Delta \vw
=
\underbrace{
\frac{(\hat{w}_q-w_q)\mH^{-1}_{q,:}}{\mH^{-1}_{qq}}
}_{\text{basic GPTQ term}}
+
\underbrace{
\balpha \vr\hat{\mX}^{\top}\mH^{-1}_{-q}
}_{\text{scaled residual term}},
\label{eq:scaled_residual_closed_form}
\end{equation}
where
$\mH^{-1}_{-q}
=
\mH^{-1}
-
\mH^{-1}_{:,q}\mH^{-1}_{q,:}/\mH^{-1}_{qq}$
denotes the inverse Hessian after eliminating the $q$-th coordinate.
The first term of this update is the basic GPTQ reconstruction update, while the second term corresponds to the residual correction whose contribution is accurately controlled by $\balpha$.
When $\balpha=0$, this update reduces to the basic GPTQ-style reconstruction update, whereas $\balpha=1$ recovers the standard GPTAQ-style residual reconstruction update.
Therefore, the scaled-residual formulation provides a unified reconstruction form that accurately controls the strength of residual correction.

\subsection{Module-Specific Residual Scaling}
\label{sec:module_specific_scaling}

The reconstruction process is performed sequentially over modules, such that the quantized output of preceding modules serves as the input to the current module.
Let $m\in\{1,\ldots,M\}$ denote the index of a reconstructed module, where $M$ is the total number of modules considered in the model.
For the $m$-th module, we denote its FP weight and reconstructed quantized weight by $\mW_m$ and $\hat{\mW}_m$, respectively, and denote the corresponding FP and quantized-path activations by $\mX_m$ and $\hat{\mX}_m$.

For a residual scaling coefficient $\balpha_m$, the scaled-residual reconstruction can be abstracted as a module-wise reconstruction operator $\mathcal{R}_m$:
\begin{equation}
\begin{aligned}
\hat{\mW}_m
&=
\mathcal{R}_m
\left(
\mW_m,\hat{\mX}_{m-1};\balpha_m
\right),\\
\hat{\mX}_m
&=
\hat{\mW}_m\hat{\mX}_{m-1},
\end{aligned}
\label{eq:module_reconstruction_process}
\end{equation}
where $\mathcal{R}_m(\cdot;\balpha_m)$ denotes the residual reconstruction process for the $m$-th module with residual coefficient $\balpha_m$.
Since $\hat{\mX}_{m-1}$, the propagated residual, and the reconstruction state vary across modules, the same residual coefficient may lead to different reconstruction responses at different modules.
Therefore, rather than adopting a single global coefficient, we seek an individual residual scaling coefficient for each module.

The MSE objective in Eq.~\eqref{eq:residual_hessian_to_mse} is derived through HA, making it less suitable as the feedback objective for estimating $\balpha_m$. We therefore adopt an HA-independent output discrepancy $D(\mX_m,\hat{\mX}_m)$, instantiated as cosine distance, to evaluate the reconstruction response under different residual coefficients.
Specifically, for module $m$, we define the desired coefficient by minimizing its module-level reconstruction objective:
\begin{equation}
\balpha_m^{\star}
=
\arg\min_{\balpha_m}
J_m(\balpha_m)
=
\arg\min_{\balpha_m}
D\!\left(
\mX_m,\hat{\mX}_m
\right),
\label{eq:module_specific_alpha}
\end{equation}
where $J_m(\balpha_m)=D(\mX_m,\hat{\mX}_m)$ denotes the module-level reconstruction objective,
$\mX_m=\mW_m\mX_{m-1}$ denotes the FP output of the $m$-th module, and $D(\cdot,\cdot)$ denotes
a module-level output discrepancy measure.
Therefore, $\balpha_m^{\star}$ represents the residual scaling coefficient
that minimizes the output discrepancy between the reconstructed quantized
module and its corresponding FP counterpart.

\subsection{PID-Based Adaptive Estimation}
\label{sec:pid_adaptive_estimation}

Eq.~\eqref{eq:module_specific_alpha} formulates the selection of $\balpha_m$ as a module-wise optimization problem that minimizes $J_m(\balpha_m)$.
However, directly obtaining $\balpha_m^{\star}$ is expensive, since exhaustive search requires repeated reconstruction over a pre-defined coefficient grid for every module, while an analytic solution is difficult due to the discrete and non-smooth nature of low-bit quantization.

Inspired by the closed-loop control principle in iterative parameter updates~\cite{an2018pid,ma2022meta}, we formulate the estimation of $\balpha_m^{\star}$ as a zero-order feedback problem and employ a classical PID controller to solve it.
At each update step, a candidate coefficient $\balpha_m^{t-1}$ is applied to the reconstruction operator in Eq.~\eqref{eq:module_reconstruction_process}, and the resulting objective value $J_m(\balpha_m^{t-1})$ defined in Eq.~\eqref{eq:module_specific_alpha} serves as the observable feedback signal.
As illustrated in Figure~\ref{fig:pid_overview}, this gradient-free iterative estimation can be compactly expressed as
\begin{equation}
\balpha_m^{t}
=
F_{\mathrm{PID}}\!\left(
J_m(\balpha_m^{t-1})
\right),
\qquad
t=1,2,\ldots,T .
\label{eq:pid_mapping}
\end{equation}
where $\balpha_m^{t}$ denotes the residual coefficient of module $m$ at the $t$-th update step, and $F_{\mathrm{PID}}(\cdot)$ denotes the PID-based feedback update that progressively estimates $\balpha_m^{\star}$ in Eq.~\eqref{eq:module_specific_alpha}.
Following the default setting in the released GPTAQ implementation, we initialize $\balpha_m^{0}=0.25$ for all modules.
The maximum number of update steps is set to $T=3$, as analyzed in Section~\ref{subsec:ablation} and Figure~\ref{fig:pid_traj}.

We adopt a PID-based update since it provides a lightweight way to stabilize zero-order coefficient estimation from reconstruction feedback.
The proportional component reacts to the current reconstruction trend, the integral component accumulates the persistent update direction, and the derivative component suppresses abrupt changes that may cause overshoot or oscillation.
Accordingly, we instantiate $F_{\mathrm{PID}}(\cdot)$ with three components: a trend-based deviation signal, an incremental PID update rule, and an early stopping criterion.

\textbf{Deviation Signal of PID.}
Since low-bit PTQ generally leaves an inherent quantization error floor~\cite{zhong2025towards}, the absolute value of $J_m(\balpha_m)$ is not a suitable zero-tracking target. We instead
adopt the finite-difference trend of the reconstruction objective with respect
to $\balpha_m$ as the feedback signal:
\begin{equation}
\begin{aligned}
g_m^{t}
&= \frac{J_m(\balpha_m^{t-1})-J_m(\balpha_m^{t-2})}
{\bigl(J_m(\balpha_m^{0})+\varepsilon_J\bigr)
\bigl(\balpha_m^{t-1}-\balpha_m^{t-2}+\varepsilon_{\balpha}\bigr)}, \\
d_m^{t} &= \tanh(-\beta g_m^{t}),
\end{aligned}
\label{eq:trend_response}
\end{equation}
where $\varepsilon_J=10^{-8}$ normalizes the objective variation, $\varepsilon_{\balpha}=10^{-6}$ stabilizes the finite-difference denominator, and $\beta=10$ controls the response sensitivity, all of which are empirically determined.
Here, $g_m^{t}$ measures whether the recent change of $\balpha_m$ reduces the reconstruction error, and $\tanh(\cdot)$ bounds the response for stable updates.

\textbf{Update Rule of PID.}
Since $J_m(\balpha_m)$ is non-smooth under low-bit quantization, directly
relying on the instantaneous trend may cause oscillation or overshoot.
We therefore adopt the incremental PID rule to aggregate multi-step feedback:
\begin{equation}
\begin{aligned}
\Delta\balpha_m^{t}
&= K_p\bigl(d_m^{t}-d_m^{t-1}\bigr)+K_i d_m^{t} \\
&\quad + K_d\bigl(d_m^{t}-2d_m^{t-1}+d_m^{t-2}\bigr),
\end{aligned}
\label{eq:pid_alpha_update}
\end{equation}
\begin{equation}
\balpha_m^{t}=\balpha_m^{t-1}+\Delta \balpha_m^{t},
\label{eq:alpha_update}
\end{equation}
where $K_p$, $K_i$, and $K_d$ are the proportional, integral, and derivative
gains, respectively, all set to $1$.
This update smooths the search trajectory and stabilizes the estimation of
the module-specific coefficient.

\begin{table*}[!t]
\centering
\caption{Compared results on large language models. We report perplexity on WikiText2 (Wiki2) and C4, together with zero-shot accuracy on six downstream tasks.}
\label{tab:llm_main}
\resizebox{\textwidth}{!}{
\begin{tabular}{l|l|l|cc|ccccccc}

\toprule
Model & W/A & Method & Wiki2 $\downarrow$ & C4 $\downarrow$ & PIQA & ARC-E & ARC-C & HellaSwag & WinoGrande & BoolQ & Avg $\uparrow$ \\
\hhline{-|-|-|--|-------}

\multirow{19}{*}{Llama2-7B}
& & FP     & 5.47 & 6.90 & 79.00 & 74.60 & 46.50 & 76.00 & 68.90 & 77.70 & 70.50 \\
\hhline{~|-|-|--|-------}
& \multirow{9}{*}{W4A4}
& RTN          & 7.99 & 11.58 & 74.16 & 62.84 & 36.35 & 66.24 & 61.17 & 69.79 & 61.76 \\
&  & GPTQ      & 6.01 & 8.16  & 77.69 & 71.76 & 42.49 & 73.17 & 65.04 & \textbf{75.32} & 67.58 \\
&  & FOEM & 6.05 & 8.22 & 76.61 & 70.54 & 42.49 & 72.49 & 65.82 & 73.43 & 66.90 \\
&  & SmoothQuant & 2.48e3 & 4.09e3 & 51.14 & 26.22 & 28.50 & 25.93 & 49.41 & 41.62 & 37.14 \\
&  & ResComp   & 5.88 & 8.04  & \textbf{77.86} & 70.46 & 42.69 & \textbf{73.72} & 65.23 & 73.06 & 67.17 \\
&  & GPTAQ     & 5.86 & 7.98  & 76.50 & 70.37 & 41.98 & 73.52 & 65.90 & 73.64 & 66.98 \\
&  & CoreQ & \textbf{5.80} & 8.02 & 77.37 & \textbf{72.73} & 43.69 & 73.06 & 67.88 & 74.89 & \textbf{68.27} \\
&  & \cellcolor{gray!20}ResComp-MARR & \cellcolor{gray!20}5.84 & \cellcolor{gray!20}8.03 & \cellcolor{gray!20}76.88 & \cellcolor{gray!20}69.91 & \cellcolor{gray!20}42.24 & \cellcolor{gray!20}73.70 & \cellcolor{gray!20}66.06 & \cellcolor{gray!20}74.71 & \cellcolor{gray!20}67.25 \\
&  & \cellcolor{gray!20}GPTAQ-MARR   & \cellcolor{gray!20}5.83 & \cellcolor{gray!20}\textbf{7.92} & \cellcolor{gray!20}77.04 & \cellcolor{gray!20}71.46 & \cellcolor{gray!20}\textbf{43.77} & \cellcolor{gray!20}73.57 & \cellcolor{gray!20}\textbf{67.96} & \cellcolor{gray!20}73.85 & \cellcolor{gray!20}67.94 \\
\hhline{~|-|-|--|-------}
& \multirow{9}{*}{W2A4}
& RTN          & 7.70e3 & 9.00e3 & 50.76 & 25.59 & \textbf{28.58} & 26.38 & 48.54 & 37.83 & 36.28 \\
&  & GPTQ      & 36.37 & 62.87 & 56.80 & 33.38 & 21.76 & 31.77 & 52.33 & 49.79 & 40.97 \\
&  & FOEM & 48.77 & 73.95 & 54.73 & 32.95 & 21.93 & 30.62 & 49.72 & 47.77 & 39.62 \\
&  & SmoothQuant & 2.17e5 & 2.59e5 & 49.13 & 25.84 & 26.62 & 26.01 & 51.54 & 45.93 & 37.51 \\
&  & ResComp   & 11.00 & 23.13 & 60.79 & \textbf{44.17} & 25.02 & 41.79 & 53.41 & 62.06 & 47.87 \\
&  & GPTAQ     & 11.24 & 19.47 & \textbf{62.13} & 44.15 & 25.68 & 42.70 & 52.64 & 58.59 & 47.65 \\
&  & CoreQ & 9.63 & 20.94 & 61.51 & 44.12 & 24.26 & 41.09 & 55.67 & \textbf{62.24} & 48.15 \\
&  & \cellcolor{gray!20}ResComp-MARR & \cellcolor{gray!20}9.92 & \cellcolor{gray!20}21.85 & \cellcolor{gray!20}60.94 & \cellcolor{gray!20}43.56 & \cellcolor{gray!20}25.17 & \cellcolor{gray!20}42.81 & \cellcolor{gray!20}55.49 & \cellcolor{gray!20}62.05 & \cellcolor{gray!20}48.34 \\
&  & \cellcolor{gray!20}GPTAQ-MARR   & \cellcolor{gray!20}\textbf{9.56} & \cellcolor{gray!20}\textbf{15.53} & \cellcolor{gray!20}61.48 & \cellcolor{gray!20}43.90 & \cellcolor{gray!20}25.26 & \cellcolor{gray!20}\textbf{43.84} & \cellcolor{gray!20}\textbf{57.38} & \cellcolor{gray!20}61.31 & \cellcolor{gray!20}\textbf{48.86} \\
\hhline{-|-|-|--|-------}

\multirow{19}{*}{Llama2-13B}
& & FP     & 4.88 & 6.41 & 80.50 & 77.50 & 49.20 & 79.40 & 72.40 & 80.60 & 73.30 \\
\hhline{~|-|-|--|-------}
& \multirow{9}{*}{W4A4}
& RTN          & 5.93 & 8.34 & 77.37 & 74.03 & 44.80 & 73.73 & 68.82 & 76.09 & 69.14 \\
&  & GPTQ      & 5.29 & 7.37 & 78.24 & 74.83 & 46.67 & 76.90 & 69.53 & 79.14 & 70.88 \\
&  & FOEM & 5.32 & 7.40 & 78.73 & \textbf{75.51} & \textbf{48.29} & 75.93 & 68.98 & 78.62 & 71.01 \\
&  & SmoothQuant & 4.72e3 & 3.94e3 & 50.82 & 26.05 & 27.99 & 26.12 & 48.30 & 38.65 & 36.32 \\
&  & ResComp   & 5.19 & 7.28 & 78.67 & 74.45 & 47.87 & 77.10 & 70.32 & \textbf{80.03} & 71.41 \\
&  & GPTAQ     & 5.18 & 7.22 & 78.24 & 73.91 & 46.84 & 77.10 & 69.46 & 78.23 & 70.63 \\
&  & CoreQ & 5.17 & 7.23 & 79.05 & 75.42 & 46.50 & 77.19 & 69.38 & 79.17 & 71.12 \\
&  & \cellcolor{gray!20}ResComp-MARR & \cellcolor{gray!20}5.16 & \cellcolor{gray!20}7.26 & \cellcolor{gray!20}\textbf{79.38} & \cellcolor{gray!20}75.17 & \cellcolor{gray!20}47.78 & \cellcolor{gray!20}77.11 & \cellcolor{gray!20}\textbf{70.96} & \cellcolor{gray!20}78.35 & \cellcolor{gray!20}\textbf{71.46} \\
&  & \cellcolor{gray!20}GPTAQ-MARR   & \cellcolor{gray!20}\textbf{5.14} & \cellcolor{gray!20}\textbf{7.17} & \cellcolor{gray!20}78.73 & \cellcolor{gray!20}73.99 & \cellcolor{gray!20}46.84 & \cellcolor{gray!20}\textbf{77.27} & \cellcolor{gray!20}70.48 & \cellcolor{gray!20}77.46 & \cellcolor{gray!20}70.79 \\
\hhline{~|-|-|--|-------}
& \multirow{9}{*}{W2A4}
& RTN          & 5.60e3 & 5.50e3 & 48.59 & 27.19 & 27.13 & 25.02 & 51.78 & 37.83 & 36.26 \\
&  & GPTQ      & 13.55 & 30.81 & 60.94 & 41.29 & 24.49 & 41.35 & 53.35 & 62.05 & 47.24 \\
&  & FOEM & 17.60 & 35.51 & 61.37 & 40.82 & 23.89 & 38.62 & 51.70 & 62.08 & 46.41 \\
&  & SmoothQuant & 2.49e5 & 2.35e5 & 49.24 & 26.73 & 30.80 & 26.15 & 48.54 & 61.71 & 40.53 \\
&  & ResComp   & 8.31 & 17.90 & 63.76 & 51.76 & 26.55 & 47.65 & 54.22 & 60.29 & 50.70 \\
&  & GPTAQ     & 8.62 & 15.61 & \textbf{67.03} & \textbf{54.83} & \textbf{31.08} & 48.54 & 56.31 & 62.01 & 53.30 \\
&  & CoreQ & 7.87 & 16.89 & 66.16 & 51.64 & 30.29 & 49.51 & \textbf{60.30} & 62.48 & 53.40 \\
&  & \cellcolor{gray!20}ResComp-MARR & \cellcolor{gray!20}7.97 & \cellcolor{gray!20}17.61 & \cellcolor{gray!20}64.74 & \cellcolor{gray!20}48.27 & \cellcolor{gray!20}28.75 & \cellcolor{gray!20}48.67 & \cellcolor{gray!20}57.62 & \cellcolor{gray!20}62.91 & \cellcolor{gray!20}51.83 \\
&  & \cellcolor{gray!20}GPTAQ-MARR   & \cellcolor{gray!20}\textbf{7.80} & \cellcolor{gray!20}\textbf{15.18} & \cellcolor{gray!20}66.38 & \cellcolor{gray!20}51.39 & \cellcolor{gray!20}30.38 & \cellcolor{gray!20}\textbf{51.93} & \cellcolor{gray!20}59.04 & \cellcolor{gray!20}\textbf{64.16} & \cellcolor{gray!20}\textbf{53.88} \\
\hhline{-|-|-|--|-------}

\multirow{19}{*}{Llama3-8B}
& & FP     & 6.44 & 9.61 & 80.70 & 77.70 & 53.70 & 79.10 & 73.20 & 81.10 & 74.30 \\
\hhline{~|-|-|--|-------}
& \multirow{9}{*}{W4A4}
& RTN          & 9.92 & 15.85 & 74.10 & 65.28 & 40.02 & 70.26 & 65.19 & 74.40 & 64.88 \\
&  & GPTQ      & 7.77 & 12.54 & 75.73 & 72.18 & 44.28 & 73.81 & 66.69 & 76.15 & 68.14 \\
&  & FOEM & 7.79 & 12.67 & 77.09 & 72.56 & 45.99 & 73.92 & 67.01 & 76.33 & 68.82 \\
&  & SmoothQuant & 2.99e3 & 2.06e3 & 50.98 & 26.35 & 24.74 & 26.40 & 52.17 & 43.43 & 37.34 \\
&  & ResComp   & 7.42 & 12.39 & 78.02 & 72.22 & \textbf{47.44} & 74.72 & 67.64 & 76.79 & 69.47 \\
&  & GPTAQ     & 7.36 & 12.22 & \textbf{78.24} & 71.80 & 46.76 & 74.94 & 67.88 & 77.86 & 69.58 \\
&  & CoreQ & 7.33 & 12.16 & 77.51 & 74.42 & 47.29 & 73.36 & 66.96 & \textbf{77.96} & 69.58 \\
&  & \cellcolor{gray!20}ResComp-MARR & \cellcolor{gray!20}7.36 & \cellcolor{gray!20}12.35 & \cellcolor{gray!20}78.18 & \cellcolor{gray!20}\textbf{74.62} & \cellcolor{gray!20}46.16 & \cellcolor{gray!20}74.97 & \cellcolor{gray!20}67.40 & \cellcolor{gray!20}76.70 & \cellcolor{gray!20}69.67 \\
&  & \cellcolor{gray!20}GPTAQ-MARR   & \cellcolor{gray!20}\textbf{7.31} & \cellcolor{gray!20}\textbf{12.15} & \cellcolor{gray!20}77.97 & \cellcolor{gray!20}73.95 & \cellcolor{gray!20}47.35 & \cellcolor{gray!20}\textbf{75.17} & \cellcolor{gray!20}\textbf{68.67} & \cellcolor{gray!20}76.97 & \cellcolor{gray!20}\textbf{70.01} \\
\hhline{~|-|-|--|-------}
& \multirow{9}{*}{W2A4}
& RTN          & 6.10e4 & 5.00e4 & 51.09 & 25.63 & \textbf{26.62} & 26.87 & 50.20 & 43.15 & 37.26 \\
&  & GPTQ      & 133.60 & 196.59 & 53.48 & 32.95 & 21.25 & 30.09 & 50.75 & 46.12 & 39.11 \\
&  & FOEM & 49.98 & 116.91 & 56.58 & 34.39 & 20.99 & 32.65 & 51.30 & 50.70 & 41.10 \\
&  & SmoothQuant & 4.07e6 & 2.18e6 & 51.14 & 23.99 & 25.34 & 26.20 & 47.43 & 47.28 & 36.90 \\
&  & ResComp   & 17.49 & 55.22 & 58.25 & 38.32 & 23.74 & 38.54 & 53.20 & 61.59 & 45.44 \\
&  & GPTAQ     & 19.33 & 34.07 & 58.70 & \textbf{40.49} & 23.21 & 38.96 & 54.22 & 62.29 & 46.31 \\
&  & CoreQ & 17.29 & 59.31 & 59.74 & 40.24 & 24.40 & 37.64 & 53.43 & \textbf{62.57} & \textbf{46.34} \\
&  & \cellcolor{gray!20}ResComp-MARR & \cellcolor{gray!20}\textbf{17.27} & \cellcolor{gray!20}52.83 & \cellcolor{gray!20}57.56 & \cellcolor{gray!20}37.37 & \cellcolor{gray!20}23.98 & \cellcolor{gray!20}38.35 & \cellcolor{gray!20}\textbf{55.25} & \cellcolor{gray!20}61.50 & \cellcolor{gray!20}45.67 \\
&  & \cellcolor{gray!20}GPTAQ-MARR   & \cellcolor{gray!20}17.43 & \cellcolor{gray!20}\textbf{31.59} & \cellcolor{gray!20}\textbf{59.79} & \cellcolor{gray!20}39.35 & \cellcolor{gray!20}25.26 & \cellcolor{gray!20}\textbf{40.58} & \cellcolor{gray!20}54.14 & \cellcolor{gray!20}58.90 & \cellcolor{gray!20}\textbf{46.34} \\
\bottomrule
\end{tabular} 
}
\end{table*}

\textbf{Stopping Criterion of PID.}
To avoid unnecessary reconstruction overhead, we stop the update when the
relative change of the reconstruction objective becomes sufficiently small:
\begin{equation}
\left|
\frac{J_m(\balpha_m^{t})-J_m(\balpha_m^{t-1})}
{J_m(\balpha_m^{0})+\varepsilon_J}
\right|<\tau,
\label{eq:alpha_stop}
\end{equation}
where $\tau$ controls the stopping tolerance for the relative objective change and we empirically set $\tau=10^{-5}$.
The current $\balpha_m^{t}$ then serves as the final residual scaling coefficient for module $m$.

With the above three components, the PID-based update in MARR efficiently
estimates a module-specific $\balpha_m$ that better balances the two-sided
effect of the residual reconstruction term with module-level reconstruction
error as feedback.

\section{Experiments} 
\label{sec:experiments}

In this section, we describe the experimental setup and results. All experiments are implemented in PyTorch and conducted on a single NVIDIA RTX A6000 GPU.
We denote each quantization setting as W$x$A$y$, where $x$ and $y$ are the bit-widths of weights and activations, respectively.

\subsection{Datasets and Experimental Settings}
\label{subsec:exp_setup}

\textbf{LLM Settings.}
Following GPTAQ~\cite{li2025gptaq}, we adopt the WikiText2~\cite{merity2016pointer} and C4~\cite{raffel2020exploring} datasets for language modeling evaluation, and the PIQA~\cite{bisk2020piqa}, ARC-Easy and ARC-Challenge~\cite{clark2018arc}, HellaSwag~\cite{zellers2019hellaswag}, WinoGrande~\cite{sakaguchi2021winogrande}, and BoolQ~\cite{clark2019boolq} datasets for zero-shot evaluation.
For reconstruction-based methods, the calibration set consists of 128 sequences of 2,048 tokens with a batch size of 8. SmoothQuant~\cite{xiao2023smoothquant} adopts 512 shuffled sentences from the Pile~\cite{gao2020pile} validation set with a 512-token cap, while AWQ~\cite{lin2024awq} employs 128 Pile-validation calibration samples with the same token cap.
We report Perplexity (PPL) for language modeling evaluation, where lower values indicate better performance, and accuracy for zero-shot evaluation, together with the average accuracy across the six zero-shot tasks.
We evaluate the quantized models with a sequence length of 2,048 and an evaluation batch size of 2. SmoothQuant adopts a smoothing factor of $0.85$, followed by per-channel weight Round-To-Nearest (RTN)~\cite{nagel2020up} and per-token activation quantization at the evaluated bit widths. AWQ employs activation-aware scale and clipping search, a group size of 128, and zero points. Its W3A16 and W2A16 results are obtained with fake quantization.

\textbf{ViT Settings.}
Following the ViT PTQ protocol~\cite{wu2025fima}, we adopt the ImageNet-1K~\cite{russakovsky2015imagenet} dataset for calibration and evaluation.
For calibration, we randomly sample 1,024 images from the training set with a batch size of 32.
All inputs are processed at a resolution of $224\times224$ with the corresponding deterministic evaluation transforms, and the evaluation batch size is set to 64.

\textbf{Experimental Details.}
We adopt the same calibration data, quantization configurations, random seeds,
and evaluation protocols for all methods.
To alleviate the influence of activation outliers on reconstruction-based
quantization, we apply the rotation-based preprocessing~\cite{ashkboos2024quarot}
to RTN and the reconstruction-based methods in the main LLM comparisons
before quantization, while SmoothQuant~\cite{xiao2023smoothquant} and AWQ~\cite{lin2024awq} apply their own scaling preprocessing.
For the reconstruction-based LLM methods, weights are asymmetrically
quantized per output channel, while activations are dynamically quantized per token along the hidden dimension with a clipping ratio of 0.9.
For our proposed MARR, the module-level feedback is measured by cosine distance.

\subsection{Weight-Activation Quantization Results on LLMs}
\label{subsec:llm_results}

To evaluate MARR under weight-activation quantization, we compare it with
RTN~\cite{nagel2020up}, the basic reconstruction method
GPTQ~\cite{frantar2022gptq}, the first-order error compensation method
FOEM~\cite{zheng2026first}, SmoothQuant~\cite{xiao2023smoothquant}
with RTN, the residual reconstruction methods GPTAQ~\cite{li2025gptaq}
and ResComp~\cite{li2026rethinking}, and the learning-free mismatch
correction method CoreQ~\cite{cha2026coreq}.
We further apply MARR to both GPTAQ and ResComp, denoted as GPTAQ-MARR and
ResComp-MARR, respectively.
Experiments are conducted on Llama2-7B, Llama2-13B~\cite{touvron2023llama}
and Llama3-8B~\cite{llama3modelcard} under W4A4 and W2A4.
Table~\ref{tab:llm_main} reports PPL on WikiText2 and C4 together
with zero-shot accuracy on the six downstream tasks described in
Section~\ref{subsec:exp_setup}.

As shown in Table~\ref{tab:llm_main}, both MARR variants lower WikiText2
and C4 PPL relative to their respective backbones in all six
model--bit-width settings. They also improve the six-task mean accuracy in
each of these settings. For example, on Llama2-7B under W2A4,
GPTAQ-MARR reduces WikiText2 PPL from 11.24 to 9.56 and C4
PPL from 19.47 to 15.53; ResComp-MARR reduces the respective
PPL values from 11.00 to 9.92 and from 23.13 to 21.85. The gains are
generally larger under W2A4 than W4A4, where a stronger accumulated
activation mismatch leaves more room for module-adaptive residual scaling.
Under the evaluated W4A4 and W2A4 SmoothQuant configurations,
PPL exceeds 2,000 on both datasets for all three models, and
the six-task mean accuracy ranges from $36.3\%$ to $40.5\%$.
Compared with CoreQ, GPTAQ-MARR achieves lower C4 PPL in all six settings
and lower WikiText2 PPL in four of them, whereas CoreQ attains the lowest
WikiText2 PPL (5.80) and the highest six-task mean accuracy ($68.3\%$) on
Llama2-7B under W4A4.
Notably, MARR consistently achieves PPL reductions of up to 20.2\% over the original residual reconstruction methods.

\begin{table*}[!t]
\centering
\caption{Compared results under weight-only quantization on large language models. We report perplexity on WikiText2 (Wiki2) and C4, together with zero-shot accuracy on six downstream tasks under W3A16 and W2A16.}
\label{tab:weight_only_main}
\resizebox{\textwidth}{!}{
\begin{tabular}{l|l|l|cc|ccccccc}
\toprule
Model & W/A & Method & Wiki2 $\downarrow$ & C4 $\downarrow$ & PIQA & ARC-E & ARC-C & HellaSwag & WinoGrande & BoolQ & Avg $\uparrow$ \\
\hhline{-|-|-|--|-------}

\multirow{21}{*}{Llama2-7B}
 &  & FP & 5.47 & 6.90 & 79.00 & 74.60 & 46.50 & 76.00 & 68.90 & 77.70 & 70.50 \\
\hhline{~|-|-|--|-------}
& \multirow{10}{*}{W3A16}
 & RTN & 146.54 & 114.65 & 53.20 & 30.60 & 24.90 & 29.10 & 48.30 & 37.80 & 37.32 \\
 &  & GPTQ & 6.10 & 8.70 & 77.10 & 71.60 & 43.30 & 71.10 & 67.00 & \textbf{75.90} & 67.67 \\
 &  & FOEM & 6.28 & 8.95 & 77.31 & 72.43 & 41.89 & 70.46 & 66.93 & 75.32 & 67.39 \\
 &  & SmoothQuant & 4.79e4 & 4.90e4 & 49.24 & 27.15 & 27.65 & 26.09 & 51.30 & 39.27 & 36.78 \\
 &  & AWQ & 6.24 & 8.30 & 77.58 & 69.49 & \textbf{44.45} & \textbf{73.26} & 68.03 & 73.49 & 67.72 \\
 &  & ResComp & 5.92 & 8.44 & 76.70 & 72.00 & 42.10 & 71.90 & 68.20 & 75.20 & 67.67 \\
 &  & GPTAQ & 5.89 & 8.36 & 77.40 & 71.80 & 42.40 & 72.00 & 66.90 & 74.10 & 67.44 \\
 &  & CoreQ & 5.84 & 8.33 & \textbf{78.29} & \textbf{72.90} & 43.34 & 72.20 & 67.17 & 72.20 & 67.68 \\
 &  & \cellcolor{gray!20}ResComp-MARR & \cellcolor{gray!20}5.88 & \cellcolor{gray!20}8.37 & \cellcolor{gray!20}77.30 & \cellcolor{gray!20}72.50 & \cellcolor{gray!20}42.70 & \cellcolor{gray!20}71.50 & \cellcolor{gray!20}66.50 & \cellcolor{gray!20}73.10 & \cellcolor{gray!20}67.25 \\
 &  & \cellcolor{gray!20}GPTAQ-MARR & \cellcolor{gray!20}\textbf{5.83} & \cellcolor{gray!20}\textbf{8.26} & \cellcolor{gray!20}77.40 & \cellcolor{gray!20}71.50 & \cellcolor{gray!20}42.50 & \cellcolor{gray!20}72.40 & \cellcolor{gray!20}\textbf{68.80} & \cellcolor{gray!20}74.60 & \cellcolor{gray!20}\textbf{67.83} \\
\hhline{~|-|-|--|-------}
& \multirow{10}{*}{W2A16}
 & RTN & 9.80e3 & 1.20e4 & 50.50 & 25.60 & 28.20 & 26.50 & 49.20 & 37.80 & 36.30 \\
 &  & GPTQ & 21.33 & 41.76 & 58.70 & 38.80 & 23.40 & 35.40 & 52.80 & 61.60 & 45.10 \\
 &  & FOEM & 35.25 & 52.28 & 57.40 & 33.04 & 21.67 & 32.61 & 51.78 & 60.61 & 42.85 \\
 &  & SmoothQuant & 2.22e5 & 2.29e5 & 50.16 & 26.39 & 28.33 & 25.92 & 48.38 & 47.31 & 37.75 \\
 &  & AWQ & 2.23e5 & 1.69e5 & 50.49 & 25.67 & 26.37 & 26.07 & 50.36 & 37.83 & 36.13 \\
 &  & ResComp & 9.39 & 18.86 & 64.70 & 49.90 & 28.50 & 46.60 & 58.30 & 63.20 & 51.85 \\
 &  & GPTAQ & 9.36 & 18.99 & 64.80 & 49.80 & 27.60 & 46.80 & 59.20 & \textbf{64.10} & 52.03 \\
 &  & CoreQ & 8.78 & \textbf{16.74} & \textbf{67.03} & \textbf{52.40} & \textbf{28.67} & \textbf{50.18} & 59.51 & 62.84 & \textbf{53.44} \\
 &  & \cellcolor{gray!20}ResComp-MARR & \cellcolor{gray!20}8.71 & \cellcolor{gray!20}18.40 & \cellcolor{gray!20}65.70 & \cellcolor{gray!20}51.40 & \cellcolor{gray!20}25.90 & \cellcolor{gray!20}47.50 & \cellcolor{gray!20}60.20 & \cellcolor{gray!20}62.20 & \cellcolor{gray!20}52.13 \\
 &  & \cellcolor{gray!20}GPTAQ-MARR & \cellcolor{gray!20}\textbf{8.59} & \cellcolor{gray!20}16.86 & \cellcolor{gray!20}65.70 & \cellcolor{gray!20}49.60 & \cellcolor{gray!20}27.70 & \cellcolor{gray!20}47.80 & \cellcolor{gray!20}\textbf{60.50} & \cellcolor{gray!20}63.90 & \cellcolor{gray!20}52.52 \\
\hhline{-|-|-|--|-------}

\multirow{21}{*}{Llama2-13B}
 &  & FP & 4.88 & 6.41 & 80.50 & 77.50 & 49.20 & 79.40 & 72.40 & 80.60 & 73.30 \\
\hhline{~|-|-|--|-------}
& \multirow{10}{*}{W3A16}
 & RTN & 48.90 & 54.60 & 57.20 & 33.60 & 23.80 & 32.10 & 53.00 & 62.10 & 43.64 \\
 &  & GPTQ & 5.36 & 7.71 & 78.30 & 75.80 & 47.10 & 76.00 & 70.80 & \textbf{79.80} & 71.30 \\
 &  & FOEM & 5.42 & 7.78 & 78.51 & \textbf{76.43} & 48.29 & 76.05 & 72.06 & 76.79 & 71.36 \\
 &  & SmoothQuant & 1.34e4 & 1.13e4 & 49.35 & 27.31 & 29.10 & 26.24 & 49.17 & 44.25 & 37.57 \\
 &  & AWQ & 5.32 & \textbf{7.31} & \textbf{79.22} & 75.08 & \textbf{48.98} & \textbf{77.46} & \textbf{72.30} & 79.66 & \textbf{72.12} \\
 &  & ResComp & 5.25 & 7.58 & 78.40 & 76.20 & 47.20 & 76.40 & 71.80 & 76.50 & 71.09 \\
 &  & GPTAQ & 5.22 & 7.50 & 78.80 & 75.70 & 47.40 & 76.20 & 71.70 & 78.30 & 71.34 \\
 &  & CoreQ & \textbf{5.18} & 7.42 & 78.62 & 75.80 & 47.18 & 76.43 & 70.32 & 77.19 & 70.92 \\
 &  & \cellcolor{gray!20}ResComp-MARR & \cellcolor{gray!20}5.22 & \cellcolor{gray!20}7.54 & \cellcolor{gray!20}78.10 & \cellcolor{gray!20}74.90 & \cellcolor{gray!20}46.70 & \cellcolor{gray!20}76.00 & \cellcolor{gray!20}71.00 & \cellcolor{gray!20}76.70 & \cellcolor{gray!20}70.57 \\
 &  & \cellcolor{gray!20}GPTAQ-MARR & \cellcolor{gray!20}\textbf{5.18} & \cellcolor{gray!20}7.45 & \cellcolor{gray!20}78.70 & \cellcolor{gray!20}75.10 & \cellcolor{gray!20}46.90 & \cellcolor{gray!20}76.30 & \cellcolor{gray!20}71.70 & \cellcolor{gray!20}78.50 & \cellcolor{gray!20}71.19 \\
\hhline{~|-|-|--|-------}
& \multirow{10}{*}{W2A16}
 & RTN & 5.20e3 & 5.00e3 & 48.40 & 26.40 & 26.60 & 25.10 & 47.00 & 37.80 & 35.22 \\
 &  & GPTQ & 9.91 & 18.72 & 66.50 & 51.40 & 28.30 & 51.20 & 58.20 & 62.40 & 52.99 \\
 &  & FOEM & 13.26 & 26.53 & 61.97 & 42.93 & 25.09 & 42.77 & 54.46 & 62.42 & 48.27 \\
 &  & SmoothQuant & 3.50e5 & 3.74e5 & 49.84 & 25.93 & 30.29 & 26.17 & 48.86 & 62.23 & 40.55 \\
 &  & AWQ & 1.23e5 & 9.55e4 & 50.05 & 26.98 & 27.73 & 25.99 & 49.64 & 62.17 & 40.43 \\
 &  & ResComp & 7.65 & 16.36 & 69.00 & 56.40 & 32.30 & 53.90 & 61.20 & 63.70 & 56.09 \\
 &  & GPTAQ & 7.71 & 15.08 & 67.80 & 53.20 & 31.00 & 51.80 & 60.90 & 62.40 & 54.51 \\
 &  & CoreQ & \textbf{6.93} & \textbf{12.99} & \textbf{69.97} & \textbf{60.35} & \textbf{35.49} & \textbf{57.30} & 62.83 & 63.21 & \textbf{58.19} \\
 &  & \cellcolor{gray!20}ResComp-MARR & \cellcolor{gray!20}7.42 & \cellcolor{gray!20}16.22 & \cellcolor{gray!20}67.40 & \cellcolor{gray!20}53.20 & \cellcolor{gray!20}30.80 & \cellcolor{gray!20}50.50 & \cellcolor{gray!20}61.10 & \cellcolor{gray!20}62.40 & \cellcolor{gray!20}54.23 \\
 &  & \cellcolor{gray!20}GPTAQ-MARR & \cellcolor{gray!20}7.08 & \cellcolor{gray!20}13.64 & \cellcolor{gray!20}69.00 & \cellcolor{gray!20}55.80 & \cellcolor{gray!20}32.90 & \cellcolor{gray!20}54.50 & \cellcolor{gray!20}\textbf{63.40} & \cellcolor{gray!20}\textbf{63.90} & \cellcolor{gray!20}56.59 \\
\hhline{-|-|-|--|-------}

\multirow{21}{*}{Llama3-8B}
 &  & FP & 6.44 & 9.61 & 80.70 & 77.70 & 53.70 & 79.10 & 73.20 & 81.10 & 74.30 \\
\hhline{~|-|-|--|-------}
& \multirow{10}{*}{W3A16}
 & RTN & 39.60 & 50.90 & 64.90 & 40.40 & 24.20 & 45.30 & 60.90 & 50.60 & 47.72 \\
 &  & GPTQ & 7.55 & 13.08 & 77.60 & 73.50 & 46.40 & 73.80 & 71.70 & \textbf{80.80} & 70.62 \\
 &  & FOEM & 7.64 & 13.21 & \textbf{78.24} & 69.82 & 44.62 & 73.97 & 71.51 & 79.30 & 69.58 \\
 &  & SmoothQuant & 5.76e5 & 5.95e5 & 52.56 & 26.01 & 26.02 & 26.32 & 50.75 & 59.66 & 40.22 \\
 &  & AWQ & 8.19 & 12.83 & 78.02 & 73.15 & 46.67 & 74.29 & \textbf{73.09} & 78.62 & 70.64 \\
 &  & ResComp & 7.30 & 12.76 & 77.60 & 72.40 & 46.40 & 73.70 & 72.00 & 78.70 & 70.11 \\
 &  & GPTAQ & 7.28 & \textbf{12.57} & 78.10 & 73.40 & 46.70 & 73.80 & 71.90 & 77.80 & 70.30 \\
 &  & CoreQ & 7.22 & \textbf{12.57} & \textbf{78.24} & \textbf{74.54} & \textbf{46.84} & \textbf{74.61} & 71.98 & 78.84 & \textbf{70.84} \\
 &  & \cellcolor{gray!20}ResComp-MARR & \cellcolor{gray!20}7.26 & \cellcolor{gray!20}12.71 & \cellcolor{gray!20}77.20 & \cellcolor{gray!20}73.70 & \cellcolor{gray!20}46.40 & \cellcolor{gray!20}73.20 & \cellcolor{gray!20}70.60 & \cellcolor{gray!20}80.10 & \cellcolor{gray!20}70.19 \\
 &  & \cellcolor{gray!20}GPTAQ-MARR & \cellcolor{gray!20}\textbf{7.18} & \cellcolor{gray!20}\textbf{12.57} & \cellcolor{gray!20}78.20 & \cellcolor{gray!20}72.50 & \cellcolor{gray!20}46.70 & \cellcolor{gray!20}73.70 & \cellcolor{gray!20}70.90 & \cellcolor{gray!20}73.90 & \cellcolor{gray!20}69.30 \\
\hhline{~|-|-|--|-------}
& \multirow{10}{*}{W2A16}
 & RTN & 4.10e4 & 3.00e4 & 51.60 & 25.40 & 25.10 & 26.80 & 50.00 & 37.80 & 36.11 \\
 &  & GPTQ & 20.30 & 59.17 & 60.50 & 43.00 & 25.00 & 41.90 & 57.00 & 63.80 & 48.52 \\
 &  & FOEM & 23.33 & 62.37 & 60.94 & 39.69 & 23.29 & 39.31 & 52.64 & 52.94 & 44.80 \\
 &  & SmoothQuant & 3.07e6 & 1.66e6 & 51.03 & 24.54 & 25.51 & 26.44 & 49.33 & 47.95 & 37.47 \\
 &  & AWQ & 1.72e6 & 2.21e6 & 50.60 & 24.96 & 25.17 & 26.48 & 51.14 & 62.17 & 40.09 \\
 &  & ResComp & 13.82 & 43.71 & 63.00 & 46.40 & 26.80 & 46.10 & \textbf{59.40} & 62.20 & 50.67 \\
 &  & GPTAQ & 13.52 & 36.47 & \textbf{65.20} & \textbf{48.00} & 27.80 & \textbf{47.30} & 58.90 & 64.30 & \textbf{51.92} \\
 &  & CoreQ & \textbf{12.75} & 37.40 & 63.49 & 45.50 & 25.94 & 45.24 & 58.25 & 65.32 & 50.62 \\
 &  & \cellcolor{gray!20}ResComp-MARR & \cellcolor{gray!20}13.70 & \cellcolor{gray!20}42.73 & \cellcolor{gray!20}60.50 & \cellcolor{gray!20}41.60 & \cellcolor{gray!20}25.70 & \cellcolor{gray!20}43.50 & \cellcolor{gray!20}57.00 & \cellcolor{gray!20}63.50 & \cellcolor{gray!20}48.63 \\
 &  & \cellcolor{gray!20}GPTAQ-MARR & \cellcolor{gray!20}13.28 & \cellcolor{gray!20}\textbf{35.90} & \cellcolor{gray!20}61.20 & \cellcolor{gray!20}42.20 & \cellcolor{gray!20}\textbf{28.60} & \cellcolor{gray!20}45.90 & \cellcolor{gray!20}56.50 & \cellcolor{gray!20}\textbf{66.10} & \cellcolor{gray!20}50.09 \\
\bottomrule
\end{tabular}
}
\end{table*}

\subsection{Weight-Only Quantization Results on LLMs}
\label{subsec:llm_wo_results}

We further evaluate MARR under weight-only quantization, where activations
remain at 16-bit precision. Under W3A16 and W2A16, we follow the same
models, baselines, and evaluation metrics as in
Section~\ref{subsec:llm_results}, and additionally include the
weight-only quantization method AWQ~\cite{lin2024awq} for comparison.
The results are reported in Table~\ref{tab:weight_only_main}.

As shown in Table~\ref{tab:weight_only_main}, GPTAQ-MARR achieves the
lowest WikiText2 PPL in four of the six model--bit-width combinations,
including a tie with CoreQ on Llama2-13B under W3A16, while CoreQ
attains the lowest WikiText2 PPL on Llama2-13B and Llama3-8B under W2A16.
For C4, GPTAQ-MARR achieves the lowest PPL on Llama2-7B under W3A16 and
Llama3-8B under W2A16 and ties with GPTAQ and CoreQ on Llama3-8B under
W3A16. CoreQ attains the lowest C4 PPL on Llama2-7B and Llama2-13B under
W2A16, and AWQ attains the lowest C4 PPL on Llama2-13B under W3A16
($7.31$ versus $7.45$ for GPTAQ-MARR). The largest MARR gains over its
backbone appear under W2A16, including $8.2\%$/$11.2\%$ PPL
reductions over GPTAQ on Llama2-7B. This confirms that controlling the
residual contribution remains beneficial when only weights are quantized.
CoreQ achieves the highest six-task mean accuracy on Llama2-7B and
Llama2-13B under W2A16 and on Llama3-8B under W3A16. AWQ achieves the
highest mean accuracy on Llama2-13B under W3A16 ($72.1\%$), while its
PPL increases sharply under W2A16 on all three models.
However, lower PPL does not consistently translate into higher zero-shot accuracy. Under W2A16, GPTAQ-MARR improves the six-task mean accuracy over GPTAQ on Llama2-7B and Llama2-13B while reducing it on Llama3-8B, despite lower PPL on all three models in Table~\ref{tab:weight_only_main}. This difference is consistent with estimating $\balpha_m$ from module-level reconstruction feedback.

\subsection{Weight-Activation Quantization Results on Larger LLMs}
\label{subsec:llm_70b}

We further evaluate Llama2-70B~\cite{touvron2023llama} and
Llama3-70B~\cite{llama3modelcard} under W4A4 and W2A4. GPTAQ, ResComp,
and their MARR variants adopt the same QuaRot-style Hadamard rotation and
128 WikiText2 calibration sequences. Each quantized model is evaluated on
both WikiText2 and C4. On Llama2-70B under W4A4, ResComp-MARR achieves the lowest PPL on both datasets (3.68/6.04). Under W2A4, it obtains the lowest WikiText2
PPL (5.59), while GPTAQ has the lowest C4 PPL (9.97). On
Llama3-70B, GPTAQ has the lowest PPL under both bit-widths; the two
MARR variants show particularly large degradation under W2A4. These results
show that the benefits of module-level residual control do not transfer
uniformly to the largest models in the most aggressive low-bit setting.

\begin{table*}[!t]
\centering
\caption{Compared results on vision transformers. We report top-1 accuracy (\%) on ImageNet under different quantization settings, where higher values indicate better performance.}
\label{tab:vit_main}
\footnotesize
\setlength{\tabcolsep}{11.4pt}
\resizebox{\textwidth}{!}{
\begin{tabular}{l|ccc|ccc|ccc}
\toprule
\multirow{2}{*}{Method}
& \multicolumn{3}{c|}{W4A4}
& \multicolumn{3}{c|}{W3A3}
& \multicolumn{3}{c}{W2A4} \\
\hhline{~|---|---|---}
& DeiT-T & DeiT-S & DeiT-B
& DeiT-T & DeiT-S & DeiT-B
& DeiT-T & DeiT-S & DeiT-B \\
\hhline{-|---|---|---}
FP    & 72.71 & 79.85 & 81.80 & 72.71 & 79.85 & 81.80 & 72.71 & 79.85 & 81.80 \\
\hhline{-|---|---|---}
PTQ4ViT      & 53.95 & 70.81 & 78.53 & 7.69  & 24.57 & 57.32 & 0.68  & 1.15  & 27.14 \\
RepQ-ViT     & 55.31 & 70.13 & 78.37 & 0.72  & 6.43  & 18.11 & 0.17  & 0.22  & 3.01  \\
AdaLog       & 62.58 & 64.81 & 76.20 & 30.05 & 55.82 & \textbf{71.90} & 1.76  & 11.90 & 46.25 \\
GPTQ         & 64.19 & 74.77 & 78.79 & 29.26 & 43.41 & 67.35 & 30.30 & 54.43 & 67.20 \\
ResComp      & 63.25 & 75.01 & 78.27 & 29.78 & 40.51 & 67.14 & 30.49 & 55.22 & 69.64 \\
GPTAQ        & 65.16 & 76.35 & 79.62 & 32.32 & 56.41 & 68.82 & 33.58 & 59.17 & 70.79 \\
\rowcolor{gray!20}
ResComp-MARR & 63.66 & 75.05 & 78.68 & 30.10 & 41.58 & 68.07 & 30.50 & 55.48 & 70.29 \\
\rowcolor{gray!20}
GPTAQ-MARR   & \textbf{65.39} & \textbf{76.81} & \textbf{80.11} & \textbf{32.45} & \textbf{59.01} & 69.22 & \textbf{33.82} & \textbf{59.92} & \textbf{71.24} \\
\bottomrule
\end{tabular}
}
\end{table*}

\begin{table}[!t]
\centering
\caption{WikiText2 (Wiki2) and C4 perplexity of Llama2-70B and Llama3-70B under W4A4 and W2A4. All results are reported with two decimal places, and PPL above 1000 is reported in e notation. Lower is better; bold denotes the lowest value for each model, bit width, and dataset.}
\label{tab:llm_70b}
\footnotesize
\setlength{\tabcolsep}{3pt}
\begin{tabular}{ll|l|rr}
\toprule
Model & W/A & Method & Wiki2 $\downarrow$ & C4 $\downarrow$ \\
\midrule
\multirow{9}{*}{Llama2-70B}
& & FP & 3.32 & 5.71 \\
\cmidrule(lr){2-5}
& \multirow{4}{*}{W4A4} & GPTAQ & 3.69 & 6.04 \\
& & ResComp & 3.69 & 6.04 \\
& & GPTAQ-MARR & 3.69 & 6.06 \\
& & ResComp-MARR & \textbf{3.68} & \textbf{6.04} \\
\cmidrule(lr){2-5}
& \multirow{4}{*}{W2A4} & GPTAQ & 5.62 & \textbf{9.97} \\
& & ResComp & 5.65 & 10.07 \\
& & GPTAQ-MARR & 5.80 & 10.65 \\
& & ResComp-MARR & \textbf{5.59} & 10.27 \\
\midrule
\multirow{9}{*}{Llama3-70B}
& & FP & 2.85 & 7.17 \\
\cmidrule(lr){2-5}
& \multirow{4}{*}{W4A4} & GPTAQ & \textbf{6.63} & \textbf{11.93} \\
& & ResComp & 7.04 & 13.30 \\
& & GPTAQ-MARR & 7.86 & 12.48 \\
& & ResComp-MARR & 7.58 & 13.70 \\
\cmidrule(lr){2-5}
& \multirow{4}{*}{W2A4} & GPTAQ & \textbf{30.88} & \textbf{64.04} \\
& & ResComp & 39.24 & 76.20 \\
& & GPTAQ-MARR & 275.64 & 1.40e3 \\
& & ResComp-MARR & 3.42e5 & 3.68e5 \\
\bottomrule
\end{tabular}
\end{table}

\subsection{Results on ViTs}
\label{subsec:vit_results}

We further evaluate MARR on ViTs under the standard ImageNet PTQ setting~\cite{wu2025fima}.
We compare it with representative ViT-oriented PTQ methods, including PTQ4ViT~\cite{yuan2022ptq4vit}, RepQ-ViT~\cite{li2023repq}, and AdaLog~\cite{wu2024adalog}, as well as reconstruction-based methods GPTQ~\cite{frantar2022gptq}, GPTAQ~\cite{li2025gptaq}, and ResComp~\cite{li2026rethinking}.
Besides W4A4 and W2A4, we also include W3A3, since visual representations in ViTs often exhibit considerable redundancy, making sub-4-bit activation quantization a meaningful stress test for low-bit PTQ~\cite{wu2025fima}.
Table~\ref{tab:vit_main} reports the top-1 accuracy of DeiT-T, DeiT-S, and DeiT-B~\cite{touvron2021training} under W4A4, W3A3, and W2A4.

As shown in Table~\ref{tab:vit_main}, MARR achieves competitive or better performance than most baselines across different DeiT models and bit-widths.
In particular, GPTAQ-MARR improves GPTAQ in most settings, indicating that the proposed module-specific residual scaling is also effective for ViTs.
This suggests that controlling the residual contribution can better balance accumulated-error correction and residual-related HA bias beyond LLMs, demonstrating the generality of MARR across Transformer architectures.

\subsection{Ablation Study}
\label{subsec:ablation}

We conduct ablation studies to analyze the key designs of MARR: module-specific residual scaling and PID-based adaptive update.

\noindent\textbf{\textcolor{black}{Residual Coefficient Ablation.}}
We compare fixed global coefficients, module-wise grid search, and PID-based estimation on GPTAQ under W2A4, under the same calibration and evaluation settings as Table~\ref{tab:llm_main}. The global variants assign one of eight coefficients, $\balpha\in\{0.25,0.50,\ldots,2.00\}$ in steps of $0.25$, to every module. For grid search, we process the modules sequentially. Within each module, we test all 41 coefficients from $0$ to $2$ in steps of $0.05$, select the coefficient that minimizes the module-output reconstruction error, and pass the resulting quantized output to the next module. The PID variant is GPTAQ-MARR, which estimates a module-specific coefficient from reconstruction feedback.
As shown in Table~\ref{tab:global_alpha_ablation}, the best fixed global coefficient differs across models. Larger global coefficients cause sharp degradation, especially for Llama3-8B. Module-wise grid search improves on the best global setting for both models, while PID estimation achieves the lowest PPL, without evaluating all 41 grid candidates in each module.

\begin{table}[!t]
\centering
\caption{W2A4 WikiText2 perplexity on Llama2-7B and Llama3-8B for fixed global coefficients, module-wise MSE grid search, and PID-estimated coefficients. Lower is better.}
\label{tab:global_alpha_ablation}
\footnotesize
\setlength{\tabcolsep}{4pt}
\begin{tabular}{ll|cc}
\toprule
\multicolumn{2}{c|}{Method} & Llama2-7B & Llama3-8B \\
\midrule
\multirow{8}{*}{\textbf{\textcolor{black}{Global $\balpha$}}} & $\balpha=0.25$ & 11.24 & 19.33 \\
 & $\balpha=0.50$ & 9.89 & 20.12 \\
 & $\balpha=0.75$ & 10.14 & 20.83 \\
 & $\balpha=1.00$ & 10.90 & 26.36 \\
 & $\balpha=1.25$ & 13.00 & 39.56 \\
 & $\balpha=1.50$ & 16.96 & 173.42 \\
 & $\balpha=1.75$ & 41.30 & 2.71e3 \\
 & $\balpha=2.00$ & 1.42e3 & 2.42e5 \\
\midrule
\multicolumn{2}{l|}{\textbf{\textcolor{black}{Grid search}}} & 9.83 & 18.36 \\
\multicolumn{2}{l|}{\textbf{\textcolor{black}{PID estimation (Ours)}}} & \textbf{9.56} & \textbf{17.43} \\
\bottomrule
\end{tabular}
\end{table}

\begin{figure*}[!t]
  \centering
  \subfloat[Llama2-7B, Layer 3, Key Proj.\label{fig:fig4a}]{%
\begin{minipage}[t]{0.32\linewidth}
\centering
    \includegraphics[width=\linewidth]{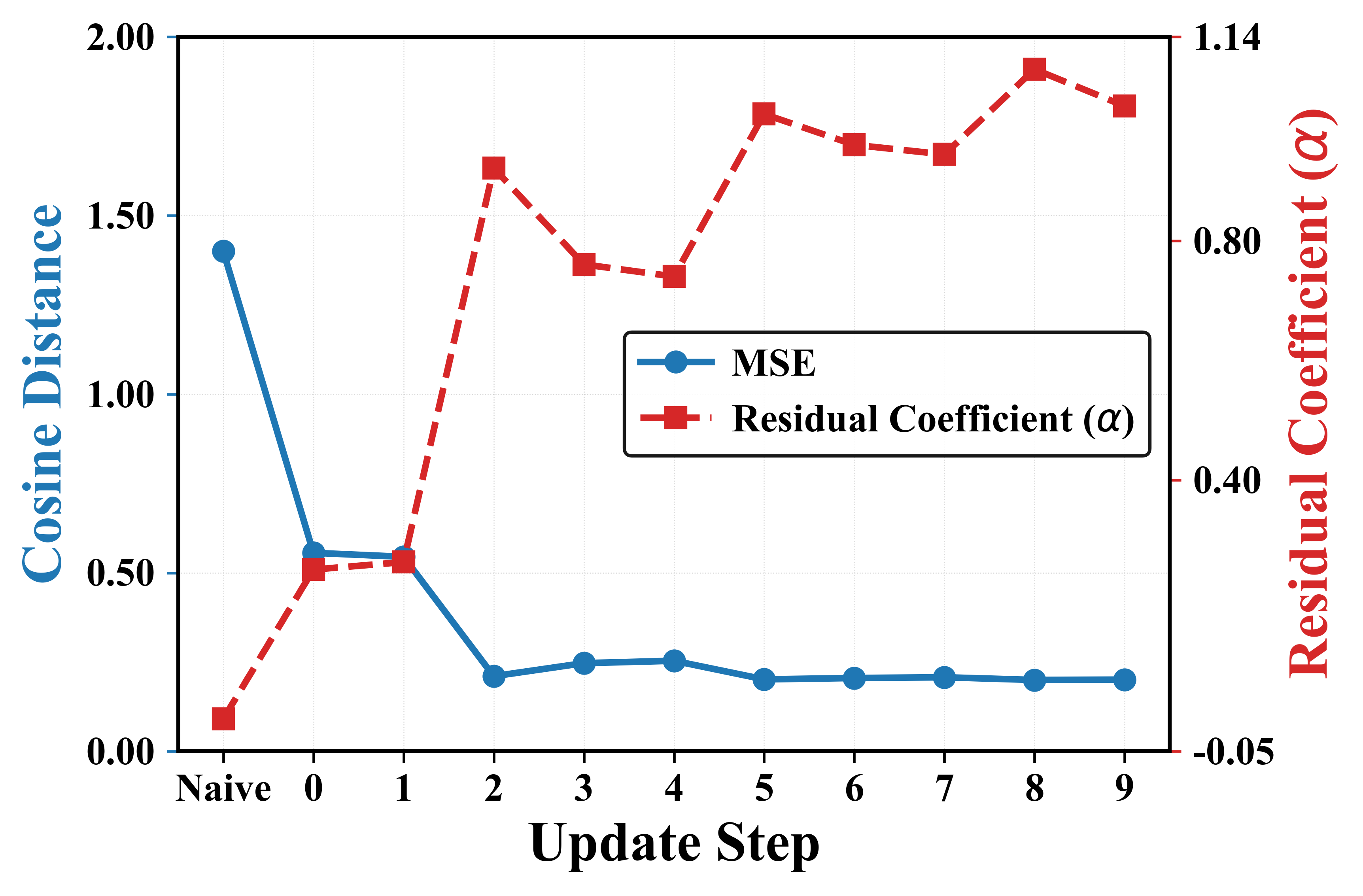}
\end{minipage}}
  \hfill
  \subfloat[Llama2-13B, Layer 13, Query Proj.\label{fig:fig4b}]{%
\begin{minipage}[t]{0.32\linewidth}
\centering
    \includegraphics[width=\linewidth]{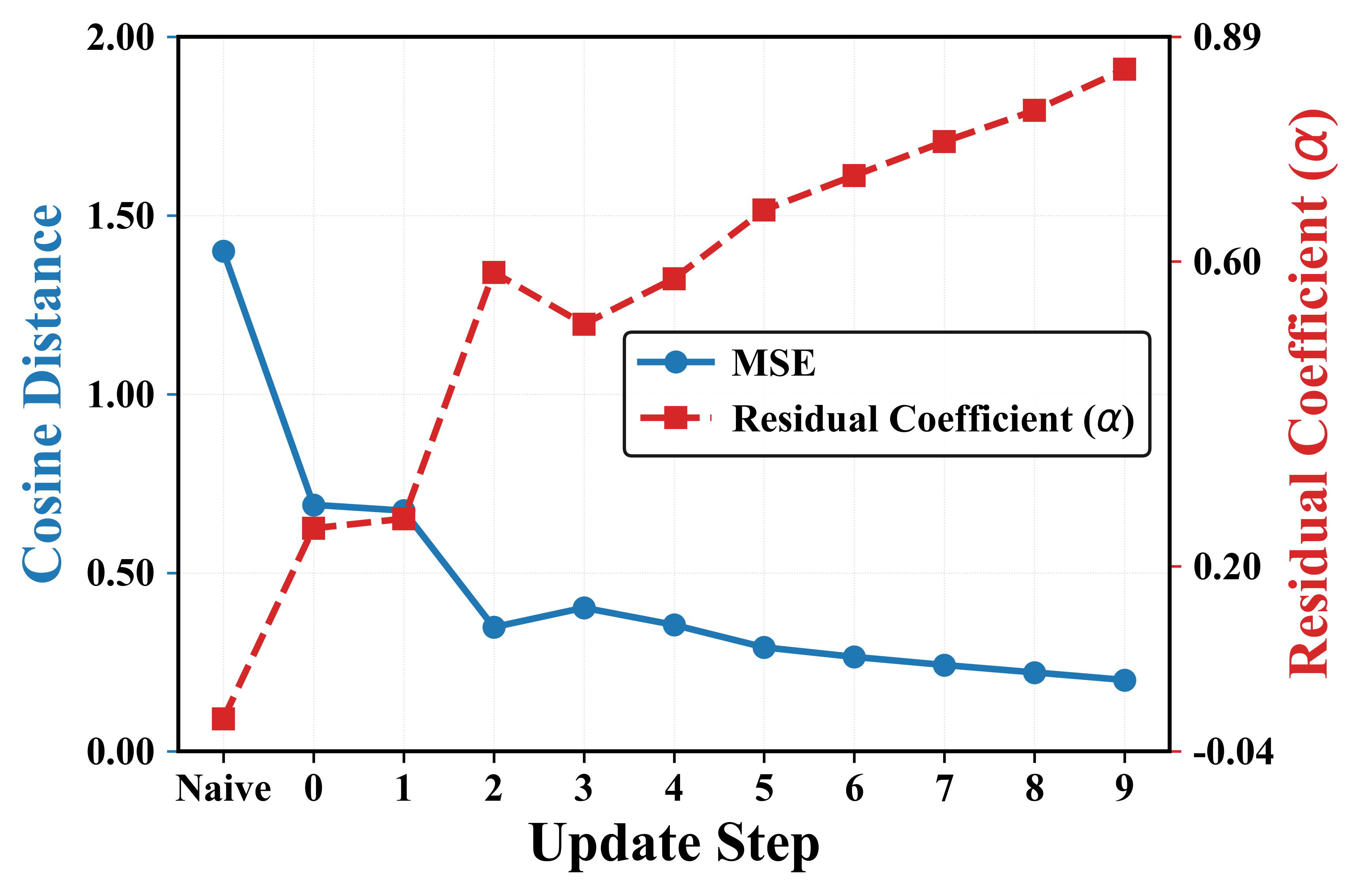}
\end{minipage}}
  \hfill
  \subfloat[Llama3-8B, Layer 12, Output Proj.\label{fig:fig4c}]{%
\begin{minipage}[t]{0.32\linewidth}
\centering
    \includegraphics[width=\linewidth]{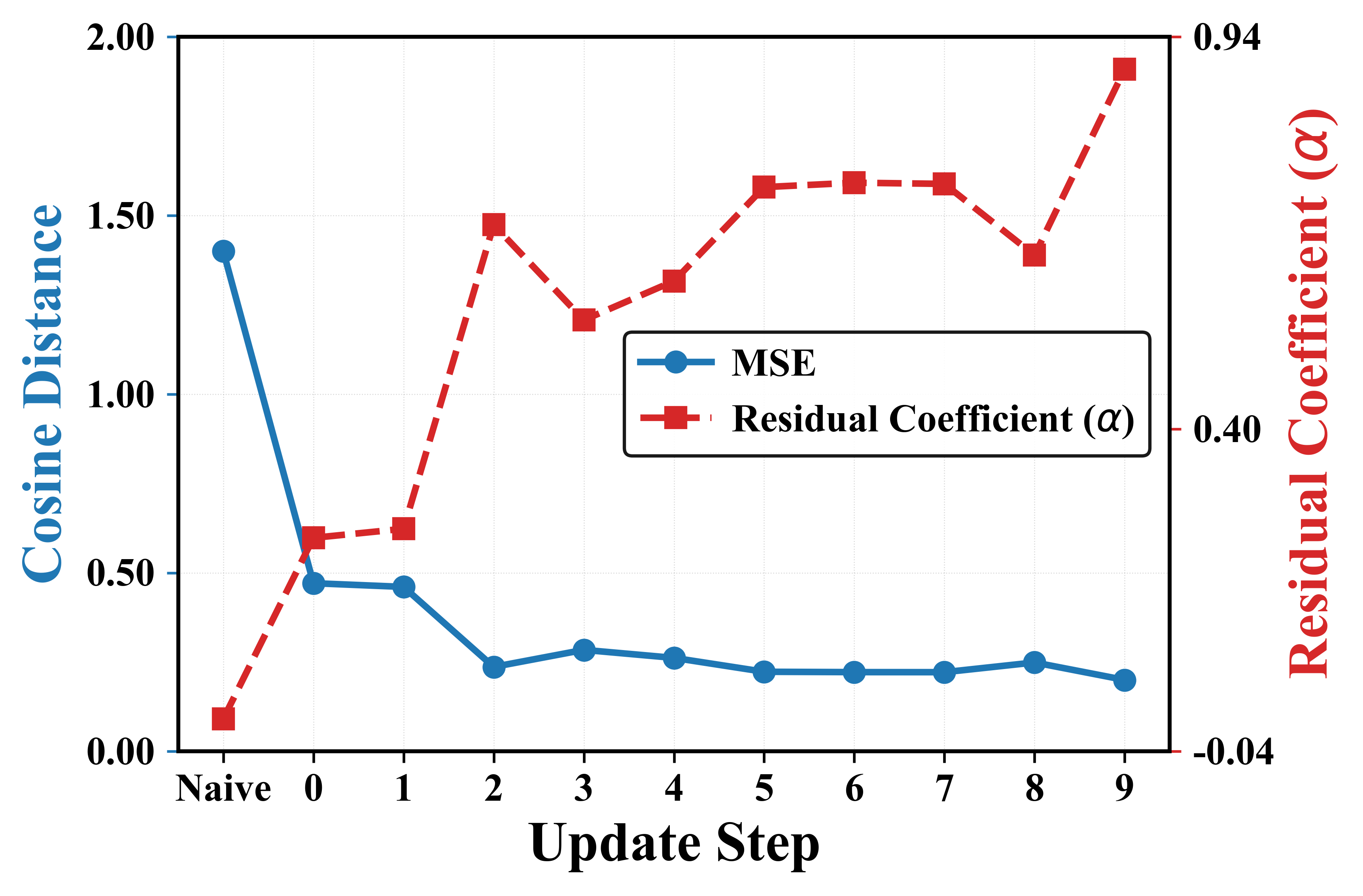}
\end{minipage}}
  \caption{
  Trajectories of the residual scaling coefficient $\balpha$ and the module-level reconstruction MSE during the PID-based adaptive update on three representative modules from Llama2-7B, Llama2-13B, and Llama3-8B under W2A4.
  }
  \label{fig:pid_traj}
\end{figure*}

\begin{figure*}[!t]
  \centering
  \includegraphics[width=\textwidth]{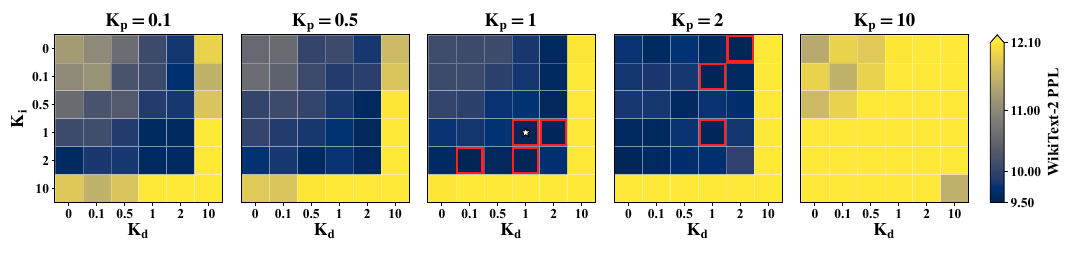}
  \caption{Sensitivity of W2A4 WikiText2 perplexity to PID gains on Llama2-7B. Each panel fixes $K_p$ and varies $K_i$ (rows) and $K_d$ (columns). The color scale is capped at 12.10 to preserve contrast among low-perplexity settings. Red boxes mark the seven settings with PPL 9.56 at two-decimal precision; the star marks the default $(K_p,K_i,K_d)=(1,1,1)$.}
  \label{fig:pid_gain_sensitivity}
\end{figure*}
  
\begin{table*}[!t]
\centering
\caption{LLM perplexity without rotation-based outlier preprocessing. Each model is evaluated under W4A4 and W2A4 on WikiText2 (Wiki2) and C4. GPTAQ-MARR and ResComp-MARR apply MARR to GPTAQ and ResComp, respectively. Lower is better; bold marks the lowest perplexity in each column.}
\label{tab:llm_no_rotation}
\footnotesize
\setlength{\tabcolsep}{7.55pt}
\resizebox{\textwidth}{!}{%
\begin{tabular}{l|cc|cc|cc|cc|cc|cc}
\toprule
\multirow{3}{*}{Method}
& \multicolumn{4}{c|}{Llama2-7B}
& \multicolumn{4}{c|}{Llama2-13B}
& \multicolumn{4}{c}{Llama3-8B} \\
\hhline{~|----|----|----}
& \multicolumn{2}{c|}{W4A4} & \multicolumn{2}{c|}{W2A4}
& \multicolumn{2}{c|}{W4A4} & \multicolumn{2}{c|}{W2A4}
& \multicolumn{2}{c|}{W4A4} & \multicolumn{2}{c}{W2A4} \\
\hhline{~|--|--|--|--|--|--}
& Wiki2 & C4 & Wiki2 & C4 & Wiki2 & C4 & Wiki2 & C4 & Wiki2 & C4 & Wiki2 & C4 \\
\hhline{-|--|--|--|--|--|--}
GPTAQ & 41.47 & \textbf{62.41} & 5.38e3 & 8.35e3 & \textbf{1.63e3} & 1.34e3 & 4.08e3 & 4.33e3 & \textbf{21.85} & 44.93 & 2.74e3 & 4.55e3 \\
ResComp & 57.13 & 65.02 & 3.21e3 & 6.67e3 & 1.96e3 & \textbf{1.13e3} & \textbf{3.77e3} & \textbf{3.78e3} & 22.64 & 52.12 & \textbf{1.69e3} & \textbf{3.48e3} \\
\rowcolor{gray!20}
GPTAQ-MARR & \textbf{33.17} & 70.92 & \textbf{3.01e3} & 3.39e3 & 3.93e3 & 5.13e3 & 6.34e3 & 6.15e3 & 22.80 & \textbf{44.60} & 2.14e3 & 9.94e3 \\
\rowcolor{gray!20}
ResComp-MARR & 34.74 & 63.48 & 3.11e3 & \textbf{2.01e3} & 4.30e3 & 4.86e3 & 5.70e3 & 6.23e3 & 22.38 & 45.32 & 4.00e3 & 1.39e4 \\
\bottomrule
\end{tabular}%
}
\end{table*}

\noindent\textbf{\textcolor{black}{PID Update Dynamics.}}
We verify the effectiveness of the PID-based adaptive update.
For visualization, we set the maximum number of update steps to 10 and record the evolution of $\balpha_m$ and module-level MSE, as shown in Figure~\ref{fig:pid_traj}. 
The results show that $\balpha_m$ progressively approaches a suitable residual strength, while the reconstruction error decreases and stabilizes within a few iterations. 
Since most modules already reach a low reconstruction error within $t=3$, we set the maximum number of update steps to $T=3$ in the main experiments. 
These results confirm that the PID-based update can efficiently estimate a module-specific residual coefficient without expensive exhaustive search.

\noindent\textbf{\textcolor{black}{PID Gain Sensitivity.}}
To investigate the sensitivity to PID gains, we evaluate Llama2-7B under W2A4 by sweeping $K_p\in\{0.1,0.5,1,2,10\}$ and $K_i,K_d\in\{0,0.1,0.5,1,2,10\}$, covering 180 configurations. As shown in Figure~\ref{fig:pid_gain_sensitivity}, moderate gains generally achieve lower WikiText2 PPL, with several configurations reaching a minimum PPL of 9.56, while increasing $K_p$ from 1 to 10 with $K_i=K_d=1$ raises PPL from 9.56 to 17.35. These results suggest that excessive gains may cause unstable feedback updates under low-bit quantization, whereas moderate gains maintain stable residual correction. Therefore, we adopt $K_p=K_i=K_d=1$ for all main experiments.

\noindent\textbf{\textcolor{black}{Impact of Rotation-Based Preprocessing.}}
To investigate the impact of rotation-based preprocessing on the effectiveness of MARR, we compare GPTAQ, ResComp, GPTAQ-MARR, and ResComp-MARR on Llama2-7B, Llama2-13B, and
Llama3-8B under W4A4 and W2A4 quantization. GPTAQ-MARR applies MARR to GPTAQ,
whereas ResComp-MARR applies it to ResComp. Unlike the main LLM experiments, this
comparison disables rotation-based preprocessing for every method. All other
quantization settings remain those in Section~\ref{subsec:exp_setup}.
Table~\ref{tab:llm_no_rotation} reports both
PPL values for every combination. Without rotation, the results depend strongly on the model and bit-width.
On Llama2-7B under W2A4, GPTAQ-MARR reduces WikiText2 and C4 PPL relative
to GPTAQ, and ResComp-MARR reduces both relative to ResComp. In contrast, on
Llama2-13B both MARR variants have higher PPL than their respective
backbones on both datasets at W4A4 and W2A4. Llama3-8B shows mixed results,
including substantially higher C4 PPL under W2A4. Thus, the rotation-free
setting does not yield consistent gains from module-adaptive residual control.

\subsection{Quantization Efficiency}
\label{subsec:efficiency}

To evaluate the quantization efficiency of MARR, Table~\ref{tab:efficiency} reports the W2A16 quantization time on a single NVIDIA RTX A6000 GPU under identical calibration and preprocessing settings. GPTAQ-MARR and ResComp-MARR require approximately $2.10$--$2.50\times$ and $2.50$--$2.70\times$ the time of GPTAQ, respectively, due to the additional PID-based coefficient estimation. Although the quantization time increases by up to nearly threefold, this overhead is incurred only once during offline PTQ and does not affect inference-time efficiency. Moreover, MARR requires at most $T=3$ update steps per module, compared with 41 candidate evaluations in exhaustive grid search, while achieving lower PPL as shown in Table~\ref{tab:global_alpha_ablation}. Therefore, MARR trades additional one-time offline quantization cost for improved quantization performance without introducing inference-time overhead.

\begin{table}[!t]
\centering
\caption{Wall-clock time of a single PTQ pass on Llama-family models under W2A16. Time is reported in minutes, and Rel. denotes the cost relative to GPTAQ, which is the residual-reconstruction backbone of GPTAQ-MARR. RTN involves no GPTQ-style iterative update and finishes in well under one minute.}
\label{tab:efficiency}
\footnotesize
\setlength{\tabcolsep}{3pt}
\begin{tabular}{l|cc|cc|cc}
\toprule
\multirow{2}{*}{Method}
& \multicolumn{2}{c|}{Llama2-7B}
& \multicolumn{2}{c|}{Llama2-13B}
& \multicolumn{2}{c}{Llama3-8B} \\
\hhline{~|--|--|--}
& Time & Rel. & Time & Rel. & Time & Rel. \\
\hhline{-|--|--|--}
RTN          & $<$1.00   & $<$0.10$\times$ & $<$1.00   & $<$0.10$\times$ & $<$1.00   & $<$0.10$\times$ \\
GPTQ         & 13.70   & 0.50$\times$    & 23.10   & 0.50$\times$    & 15.10   & 0.40$\times$    \\
ResComp      & 29.90   & 1.00$\times$    & 47.60   & 0.90$\times$    & 34.60   & 0.80$\times$    \\
GPTAQ        & 30.00   & 1.00$\times$    & 50.10   & 1.00$\times$    & 41.10   & 1.00$\times$    \\
\rowcolor{gray!20}
ResComp-MARR & 74.30   & 2.50$\times$    & 132.80  & 2.70$\times$    & 107.40  & 2.60$\times$    \\
\rowcolor{gray!20}
GPTAQ-MARR   & 72.70   & 2.40$\times$    & 125.10  & 2.50$\times$    & 88.10   & 2.10$\times$    \\
\bottomrule
\end{tabular}
\end{table}

\subsection{Limitations}
\label{subsec:limitations}

MARR has two main limitations. First, the PID-based coefficient estimation increases the quantization cost to $2.10$--$2.70\times$ that of GPTAQ (Table~\ref{tab:efficiency}). This overhead arises only once during offline PTQ, and the inference process of the quantized model remains unaffected, since MARR changes only how the quantized weights are reconstructed. Second, the advantage of MARR becomes less evident under weight-only quantization and on larger models. Under W3A16 and W2A16, the activations remain at 16-bit precision, so the smaller FP--Quant activation mismatch leaves limited room for residual regulation, and CoreQ or AWQ attains the best results in several settings (Table~\ref{tab:weight_only_main}). On Llama3-70B, both MARR variants have higher PPL than their backbones under W4A4 and W2A4 (Table~\ref{tab:llm_70b}), indicating that module-level residual control does not transfer uniformly to the largest models.

\section{Conclusion}
\label{sec:conclusion}

This paper studies residual reconstruction in low-bit PTQ from the perspective of residual contribution control.
Although activation residuals can reduce accumulated cross-module quantization errors, they may also introduce residual-related bias under the HA reconstruction objective.
To address this issue, we propose MARR, which assigns a module-specific scaling coefficient to the residual term.
By explicitly controlling the residual contribution in each module, MARR balances accumulated-error correction and residual-induced approximation bias.
We further design a PID-based adaptive update strategy that treats reconstruction error as feedback to efficiently estimate the coefficient without exhaustive search.
Experiments on LLMs and ViTs show that MARR often improves residual reconstruction baselines under low-bit quantization.

\bibliographystyle{IEEEtran}
\bibliography{refs}

\raggedbottom
\makeatletter
\expandafter\patchcmd\csname\string\IEEEbiography\endcsname{plus 1fil}{plus 1pt}{}{\PackageWarning{MARR}{IEEEbiography spacing patch failed}}
\makeatother
\begin{IEEEbiography}[{\includegraphics[width=1in,height=1.25in,clip,keepaspectratio]{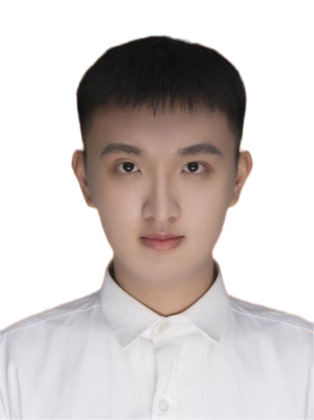}}]{Le Su}
received the B.S. degree from the School of Automation, Guangdong Polytechnic Normal University, Guangzhou, China, in 2022, and the M.S. degree from the School of Computer Science and Technology, Guangdong University of Technology, Guangzhou, China, in 2025. He is currently pursuing the Ph.D. degree with the School of Intelligent Systems Engineering, Sun Yat-sen University, Shenzhen, China. His current research interests include model quantization, efficient inference, and model compression.
\end{IEEEbiography}

\begin{IEEEbiography}[{\includegraphics[width=1in,height=1.25in,clip,keepaspectratio]{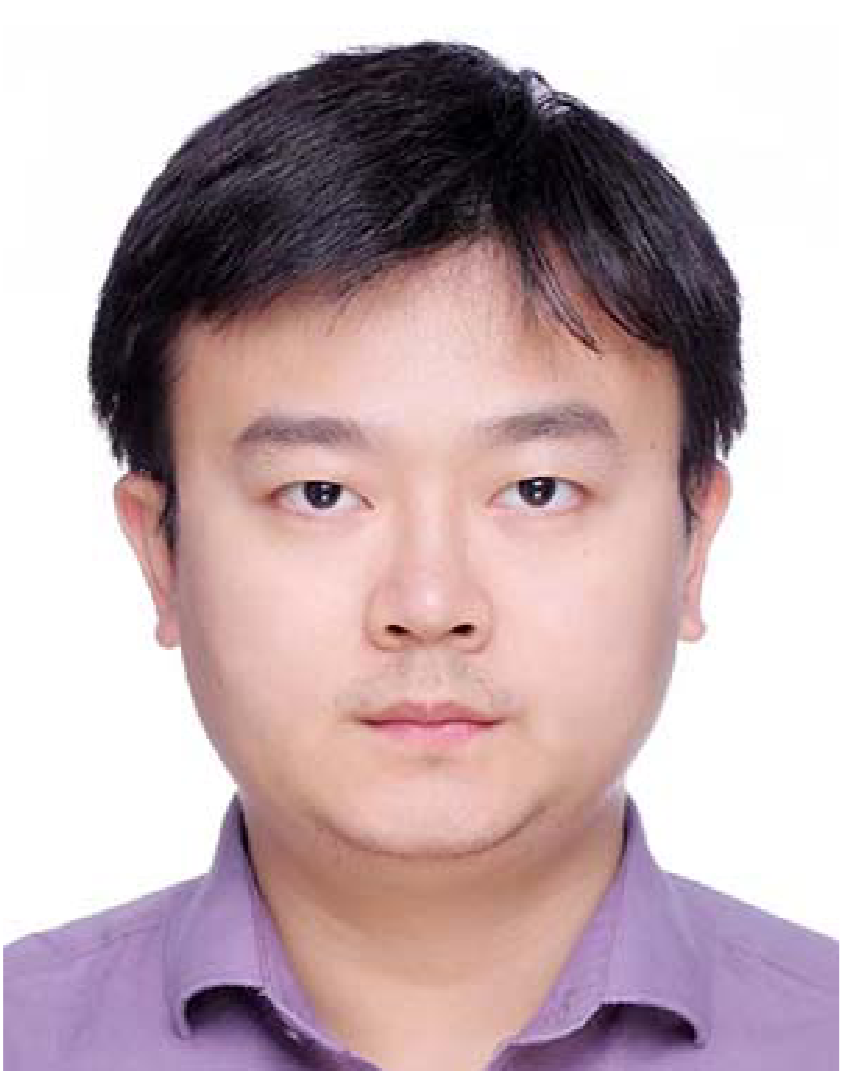}}]{Xing Luo}
received the Ph.D. degree from the University of Liverpool, Liverpool, U.K., in 2019. From 2019 to 2021, he was a Postdoctoral Researcher with Peng Cheng Laboratory, Shenzhen, China. He is currently an Assistant Research Fellow with the Department of Frontier Research, Peng Cheng Laboratory. His research interests include scientific machine learning, smart grid, and renewable energy forecasting.
\end{IEEEbiography}

\begin{IEEEbiography}[{\includegraphics[width=1in,height=1.25in,clip,keepaspectratio]{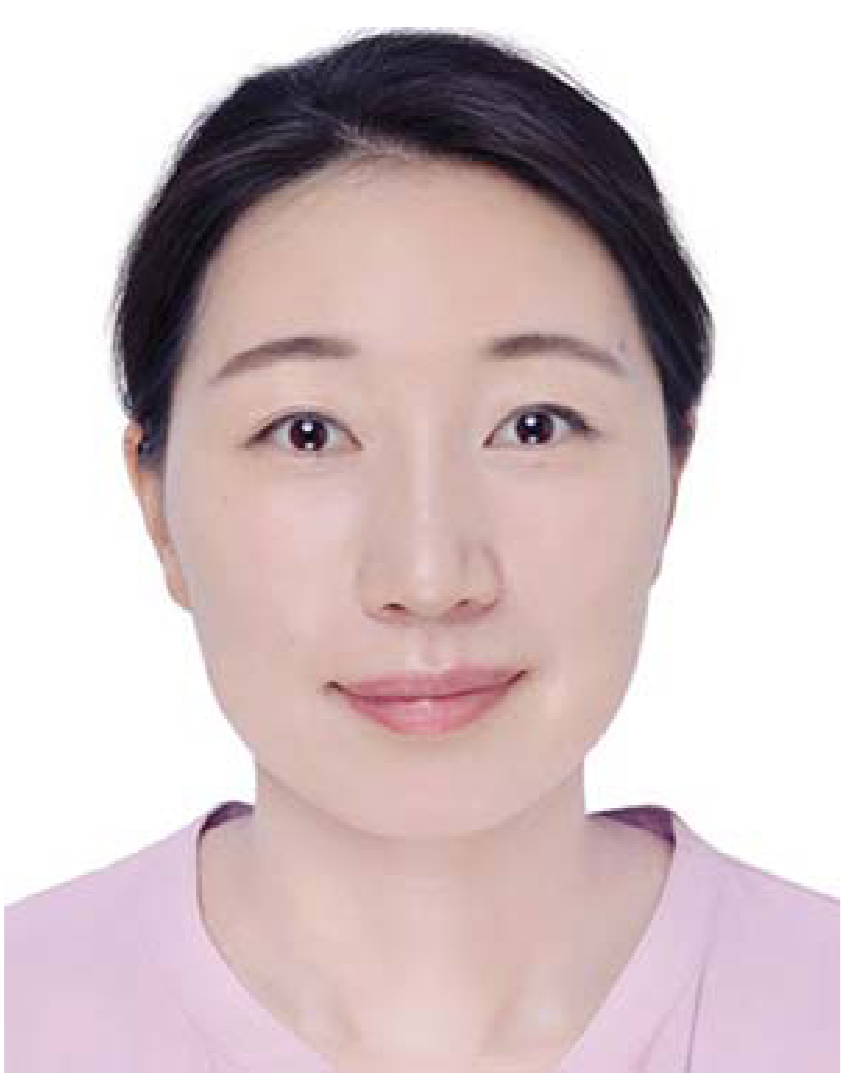}}]{Zhi Jin}
(Senior Member, IEEE) received the B.S. degree in telecommunication engineering from the University of Liverpool, Liverpool, U.K., and Xi'an Jiaotong-Liverpool University, Suzhou, China, in 2011, and the Ph.D. degree from the University of Liverpool, Liverpool, U.K., in 2016. From 2016 to 2018, she was a Postdoctoral Researcher with Shenzhen University, Shenzhen, China. During 2017--2018, she was a Research Fellow with the Technical University of Munich, Munich, Germany. She is currently an Associate Professor with the School of Intelligent Systems Engineering, Sun Yat-sen University, Shenzhen, China. Her research interests include 2D/3D image/video quality enhancement, image-based healthcare, and 3D reconstruction.
\end{IEEEbiography}

\end{document}